\documentclass{article}

\usepackage{microtype}
\usepackage{graphicx}
\usepackage{subfigure}
\usepackage{booktabs} 

\usepackage{hyperref}

\usepackage[accepted]{icml2024}

\usepackage{amsmath}
\usepackage{amssymb}
\usepackage{mathtools}
\usepackage{amsthm}

\usepackage[capitalize,noabbrev]{cleveref}

\theoremstyle{plain}
\newtheorem{theorem}{Theorem}[section]
\newtheorem{proposition}[theorem]{Proposition}
\newtheorem{lemma}[theorem]{Lemma}

\theoremstyle{definition}

\theoremstyle{remark}

\usepackage[textsize=tiny]{todonotes}
\usepackage[algo2e,ruled,vlined]{algorithm2e}
\usepackage{algorithmic}
\usepackage{multirow}

\icmltitlerunning{Bayesian Self-Supervised Contrastive Learning}

\begin{document}

\twocolumn[
\icmltitle{Bayesian Self-Supervised Contrastive Learning}




\begin{icmlauthorlist}
\icmlauthor{Bin Liu}{sch,yyy}
\icmlauthor{Bang Wang}{yyy}
\icmlauthor{Tianrui Li}{sch}
\end{icmlauthorlist}

\icmlaffiliation{sch}{School of Computing and Artificial Intelligence, Southwest Jiaotong University, Chendu, China}
\icmlaffiliation{yyy}{School of Electronic Information and Communication, Huazhong University of Science and Technology, Wuhan, China}

\icmlcorrespondingauthor{Bang Wang}{wangbang@hust.edu.cn}
\icmlcorrespondingauthor{Tianrui Li}{trli@swjtu.edu.cn}

\icmlkeywords{Machine Learning, ICML}

\vskip 0.3in
]



\printAffiliationsAndNotice{} 

\begin{abstract}
Contrastive learning has witnessed numerous successful applications across diverse domains in recent years. However, its self-supervised version still presents several exciting challenges. One prominent challenge arises from the semantic inconsistency of negative samples between the pretext task and the downstream tasks. This paper addresses this challenge by debiasing false negative and mining hard negative  within a Bayesian framework to align the semantics of negative samples across tasks. Consequently,  a modified self-supervised contrastive loss, termed Bayesian Contrastive Learning (BCL) loss, is proposed. BCL offers a flexible and principled approach to self-supervised contrastive learning and presents a generalized perspective of contrastive loss. We relate the weights used for correcting the contrastive loss and the posterior probability, and analyzed the consistency of BCL as an estimation of the contrastive loss within the supervised setting. Experimental validation confirms the effectiveness of the BCL loss.
\end{abstract}

\section{Introduction}\label{Sec:Intro}
Learning good representations without supervision has been a long-standing problem in machine learning~\cite{Arora:2019:ICML,He:2020:CVPR,Chu:2023:PAMI}. Many state-of-the-art models utilize self-supervised learning (SSL) techniques~\cite{Chen:2020:NIPS,BYOL:2020:NIPS,Liu:2021:TKDE,tong2023emp} to design pretext tasks for (pre)training the models. For instance, generative methods train models to reconstruct input data~\cite{kingma2018glow,mikolov2013distributed}, while contrastive methods train models to encode differential features between positive and negative samples\cite{chopra2005learning,hadsell2006dimensionality,gidaris2018unsupervised,Hjelm:2018:Arxiv,Chen:2020:ICML,BYOL:2020:NIPS,He:2020:CVPR,Wang:2020:ICML,pmlr-v139-radford21a}. Self-supervised learning has been extensively researched for its advantages in learning representations without the need for human labelers to manually label the data~\cite{He:2020:CVPR,Tian:2020:ECCV,Chen:2020:ICML,radford2021learning,wu2023understanding,luo2023segclip}. 

\begin{figure}[!]
	\centering
	\includegraphics[width=0.45\textwidth]{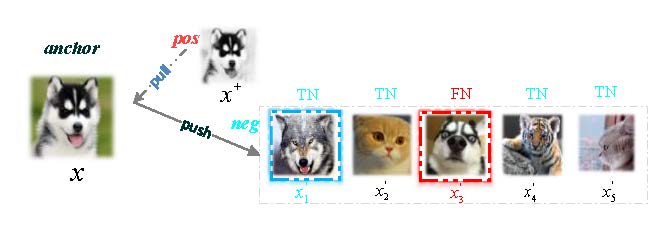}
	\caption{False negative samples and hard negative samples.}
	\label{Fig:illustrative}
\end{figure}

Contrastive learning is a primary implementation form of self-supervised learning, and remarkable successes have been observed for many applications in different domain~\cite{Alec:2019,misra:2020:CVPR,He:2020:CVPR,Tian:2020:ECCV,Chen:2020:NIPS,Liu:2021:TKDE}. However, self-supervised learning derives ``pseudo-labels" from co-occurring inputs to relate information~\cite{Liu:2021:TKDE}. For negative samples, there exists a distinction in the semantic interpretation of ``classes" in pre-training phase and the generalized semantic of ``classes" used in downstream tasks.
As illustrated in Fig.~\ref{Fig:illustrative}, for the anchor point ``dog", the semantic of negative samples is all the unlabeled samples excluding the anchor data point itself in the pre-training phase. That is, each sample is treated as an individual class. However, in the downstream task, the semantic of negative sample are samples labeled as ``not dogs". 

\par
Aligning the semantic representation of negative examples between the pretraining tasks and downstream tasks is crucial for improving performance in the latter. In the pretraining phase, there are two types of negative samples that disrupt the semantic structure of generalized ``classes". The first type consists of false negative (FN) samples, represented by $x_3^\prime$ in Fig.\ref{Fig:illustrative}. Despite being labeled as ``dog", it were erroneously treated as negative belonging to a different class than the anchor data point. This necessitates the task of false negative debiasing, aiming to prevent the false negative samples from being erroneously pushed apart. The second type comprises hard negative (HN) samples, exemplified by $x_1^\prime$ in Fig.~\ref{Fig:illustrative}. These samples possess a ``wolf" label and exhibit similarities to the anchor data point, despite belonging to a distinct class. It is essential to push these samples further apart, corresponding to the task of hard negative mining, otherwise the underlying semantic structure can be disrupted~\cite{Feng:2021:CVPR,Chuang:2020:NIPS}.

In this work, we focus on the end-to-end self-supervised contrastive learning method with SimCLR as a representative approach. We address two critical tasks in self-supervised contrastive learning, namely debiasing false negatives and mining hard negatives, within a Bayesian framework. These tasks are effectively achieved through re-weighting the negative samples, thereby correcting the contrastive loss within the self-supervised setting and obtaining the modified Bayesian self-supervised contrastive loss (BCL), which offers a flexible and principled framework for self-supervised contrastive learning and presents a generalized perspective of contrastive loss. We analyzed the small-sample properties of BCL, and established the relationship between the importance weight applied to each unlabeled sample and the posterior probability of them being true negatives. Furthermore, we investigated the large-sample properties of BCL and demonstrated its consistency as an estimation of the contrastive loss under the supervised setting.

\section{Related Work}
\par
Contrastive learning adopts the \textit{learn-to-compare} paradigm~\cite{Gutmann:2010:ICAIS} that discriminates the observed data from noise data to relieve the model from reconstructing pixel-level information of data~\cite{Oord:2018:arxiv}. Although the representation encoder $f$ and similarity measure vary from task to task~\cite{Devlin:2018:bert,He:2020:CVPR,Dosovitskiy:2014:NIPS}, they share an identical basic idea of contrasting similar pairs $(x, x^+)$ and dissimilar pairs $(x, x^-)$ to train $f$ by optimizing a contrastive loss~\cite{Wang:2020:ICML,Gutmann:2010:ICAIS,Oord:2018:arxiv,Hjelm:2018:Arxiv,Wang:2020:ICML}. The most commonly used one is the InfoNCE\cite{Oord:2018:arxiv} loss, which is formulated as follows:
\begin{tiny}
	\begin{equation}\label{Eq:SupLoss}
		\mathcal{L}_\textsc{Sup} = \mathbb E_{\substack{x \sim p_d\\~x^+ \sim p^+\\{~x^- \sim p^-}} }
		[-\log \frac{e^{f(x)^Tf(x^+)}}{e^{f(x)^Tf(x^+)} + \sum_{i=1}^Ne^{f(x)^Tf(x_i^-)}}]
	\end{equation}
\end{tiny}
\par
In the equation, $f: \mathcal{X} \rightarrow \mathbb{R}^d/t$ represents an encoder that maps a data point $x$ to the hypersphere $\mathbb{R}^d$ with a radius of $1/t$, where $t$ is the temperature scaling. For the purpose of analysis, we set $t=1$. The term $p^+$ denotes the class conditional distribution of positive samples that have the same label as $x$, while $p^-$ represents the class conditional distribution of negative samples that have a different label from $x$. 

Contrastive loss implicitly~\cite{Oord:2018:arxiv} or explicitly~\cite{Hjelm:2018:Arxiv} lower bounds the mutual information. Arora et al.~\cite{Saunshi:2019:ICML} provide theoretical analysis on the generalization bound of contrastive learning for classification tasks. Remarkable successes of supervised contrastive learning have been observed for many applications in different domains~\cite{Henaff:2020:ICML,Khosla:2020:NIPS}, but the great limitations are from their dependent on manually labeled datasets~\cite{Liu:2021:TKDE}. Self-supervised contrastive learning~\cite{Chen:2020:ICML,Chen:2020:NIPS,He:2020:CVPR,Henaff:2020:ICML,Xu:2022:Arxiv} has been extensively researched for its advantages of learning representations without human labelers for manually labeling data, and benefits almost all types of downstream tasks~\cite{Liu:2021:TKDE,Bachman:2019:NIPS,chen2020improved,Huang:2019:ICML,Wu:2018:CVPR,Zhuang:2019:CVPR}. The common practice is to obtain positive sample $x^+$ from some semantic-invariant operation on an anchor $x$ with heavy data augmentation~\cite{Chen:2020:ICML}, like random cropping and flipping~\cite{Oord:2018:arxiv}, image rotation~\cite{gidaris2018unsupervised}, cutout~\cite{DeVries:2017:arXiv} and color distortion~\cite{Szegedy:2015:CVPR}; While drawing negative samples $x^-$s simply from un-labeled data, which introduces the false negative problem, leading to incorrect encoder training. This problem is related to the classic positive-unlabeled (PU) learning~\cite{Jessa:2020:ML}, where the unlabeled data used as negative samples would be down-weighted appropriately~\cite{Du:2015:ICML,Du:2014:NIPS,Kiryo:2017:NIPS}. However, existing PU estimators are not directly applicable to contrastive loss. 

In the context of \textit{self-supervised contrastive learning}, as samples' labels are not available, the standard approach is to draw $N$ samples from the data distribution $p_d$, which are supposed to be negative samples to $x$, to optimize the following InfoNCE \textit{self-supervised contrastive loss}:
\begin{tiny}
	\begin{equation}\label{Eq:BiasLoss}
		\small
		\mathcal{L}_\textsc{Bias}= \mathbb E_{ \substack{x \sim p_d\\~x^+ \sim p^+\\ {~x^\prime \sim p_d}}}
		[-\log \frac{e^{f(x)^Tf(x^+)}}{e^{f(x)^Tf(x^+)} + \sum_{i=1}^{N}e^{f(x)^Tf(x_i^\prime)}}].
	\end{equation}
\end{tiny}
\par
Following the DCL~\cite{Chuang:2020:NIPS}, it is also called as \textit{biased contrastive loss} since those supposedly negative samples $x^\prime$ drawn from unlabeled data distribution $p_d$.  The sample $x^\prime$ could be potentially with the same label as $x$, i.e., it is a \textit{false negative} to $x$. In such a case, the false negative sample, or called sampling bias~\cite{Chuang:2020:NIPS}, would degrade the learned representations~\cite{Feng:2021:CVPR,Chuang:2020:NIPS}. Detailed sampling bias analysis is presented Appendix~\ref{sec:bias}.


\section{The Proposed Method}
\subsection{False Negative and Hard Negative} \label{sec:31}
We utilize Fig.~\ref{Fig:illustrative} to illustrate the two most crucial tasks in self-supervised contrastive learning: false negative debiasing and hard negative mining. Let's consider an anchor point labeled as ``dog." Positive samples are obtained through the application of semantically invariant image augmentation, while negative examples are randomly sampled from unlabeled images. Let us assume that we obtain five random negative samples, denoted as ${x_1^\prime,x_2^\prime,x_3^\prime,x_4^\prime,x_5^\prime}$. Among them, $x_3^\prime$ has a true label of "dog," making it a false negative sample (FN) belonging to the same class as the anchor point. The remaining four random negative samples are true negative samples (TN) that belong to different classes than the anchor point. Among the TN samples, $x_1^\prime$ corresponds to the class "wolf"; however, it bears a visual resemblance to the anchor point, thus qualifying as a hard negative sample. 

\textit{False negative debiasing}: For false negative example $x_3^\prime$, which belongs to the same class as the anchor point, we should prevent it from being pushed away from the anchor point. Instead, we aim to keep samples of the "dog" class close to each other (alignment). \textit{Hard negative mining}: For true negative example $x_1^\prime$, which belongs to a different class from the anchor point, we should push it further away. This helps maintain a larger distance between samples of the "dog" and "wolf" classes (uniformity).

Inspired by the classical technique of importance sampling, we introduce an importance weight $\omega_i$ for each unlabeled sample $x_i^\prime$ to align the semantic of negative samples. Consequently, the contrastive loss function incorporating the importance weights can be rewritten as follows:
\begin{small}
	\begin{eqnarray}
		\mathcal{L} &=& \mathbb E_{ \substack{x \sim p_d\\~x^+ \sim p^+\\{~x^\prime \sim p_d}}}
		[-\log \frac{e^{f(x)^Tf(x^+)}}{e^{f(x)^Tf(x^+)} + \sum_{i=1}^{N} \omega_i \cdot e^{f(x)^Tf(x_i^\prime)}}] \nonumber \\
		&=& \mathbb E_{ \substack{x \sim p_d\\~x^+ \sim p^+\\{~x^\prime \sim p_d}}}
		[-\log \frac{\hat{x}^+}{{\hat{x}^+} + \sum_{i=1}^{N} \omega_i \cdot \hat{x}_i}], \label{eq:debias}
	\end{eqnarray}
\end{small}

where $\hat{x}^+$ represents the similarity score between the positive example and the anchor point, given by $e^{f(x)^Tf(x^+)}$. This notation is chosen to maintain simplicity in the analysis. Next, we investigate the impact of the importance weights $\omega_i$ on the learning and updating of the model's parameters with respect to the unlabeled samples $x^\prime_i$. When incorporating importance weights, the gradient of the loss function with respect to the model parameters $\Theta$ can be computed as follows:
\begin{eqnarray}
	\frac{\partial\mathcal{L}}{\partial \Theta}
	&=&  \frac{ \omega_i\hat{x}_i} {\hat{x}^+ + \sum_{i=1}^{N} \omega_i \cdot \hat{x}_i}\cdot \frac{\partial \hat{x}_i}{\partial\Theta}. \label{eq:rule}
\end{eqnarray}
The first term, $\frac{\omega_i\hat{x}_i}{\hat{x}^+ + \sum_{i=1}^{N} \omega_i \cdot \hat{x}_i}$, is determined by the loss function, while the second term, $\frac{\partial \hat{x}_i}{\partial\Theta}$, is determined by the encoder and the scoring function. It can be observed that the importance weight $\omega_i$ appears in the numerator of the first term, acting as a modifier. It directly controls the contribution of the unlabeled sample $x_i^\prime$ to the model's parameter updates, thereby influencing how far these samples are pushed apart.

Reasonable weights $\omega_i$ should satisfy the following criteria: If the unlabeled sample $x_i^\prime$ is a false negative, reasonable weight $\omega_i$ should be small, preferably zero, to prevent the positive example from being pushed apart and maintain alignment among positive samples. This reflects the principle of \textit{false negative debiasing}. False negatives can be considered as noise present in the negative samples, and small $\omega_i$ values can cause the gradient in Eq.~\eqref{eq:rule} to vanish, preventing the model from learning patterns from noisy samples. If the unlabeled sample $x_i^\prime$ is a true negative example, a reasonable weight $\omega_i$ should be large to push hard negative samples apart and maintain uniformity among negative samples. This reflects the principle of \textit{hard negative mining}. Larger $\omega_i$ values can increase the magnitude of the gradient in Eq.~\eqref{eq:rule}, encouraging the model to learn decision boundaries between hard negative samples. The key problem is how to calculate reasonable weights that satisfy the aforementioned criteria.

\subsection{Calculation of Importance Weights}
According to the prescribed steps of Monte Carlo importance sampling \cite{Hesterberg:1988,Bengio:2008:TNN}, the calculation of importance weights requires the distribution of \textit{target sampling population} and the distribution of \textit{actual sampling population}. The importance weight is then obtained as the density ratio of the target sampling population and the actual sampling population. To derive this weight, we first focus on the similarity score $\hat{x}$ between the unlabeled sample and the anchor point. A more general formulation of the similarity score is given by $\exp(\text{sim}(f(x),f(x^\prime))/t)$. Therefore, the distribution of $\hat{x}$ is influenced by various factors, including the choice of the anchor point $x$, the architecture of the neural network $f(\cdot)$, and the specific form of the score function $\text{sim}(\cdot,\cdot)$. 

In existing research, parameterized distributions like Beta distribution \cite{Xia:2022:ICML} or the von Mises-Fisher (VMF) distribution \cite{Robinson:2021:ICLR,dong2023synthetic} are introduced. However, this assumption is often too strong. We do not make any parameterized assumptions about the distribution of $\hat{x}$, and calculate the importance weight analytically. This approach, referred to as non-parametric approach in statistical theory, sets us apart from existing methods.

\subsubsection{Target Sampling Population}
Assume that $\hat{x}$ is independently and identically distributed with $\phi$, with cumulative distribution function $\Phi(\hat{x}) = \int_{-\infty}^{\hat{x}} \phi (t) dt$. As $x^\prime$ can be either a $\textsc{Tn}$ sample or a $\textsc{Fn}$ sample, so $\phi$ contains two populations, denoted as $ \phi_{\textsc{Tn}}$ and  $\phi_{\textsc{Fn}}$. Then class conditional distribution of true negative $ \phi_{\textsc{Tn}}$ is the target sampling population.

Consider $n$ random variables from $\phi$ arranged in the ascending order according to their realizations. We write them as $X_{(1)} \le X_{(2)} \le \cdots \le X_{(n)} $, and $X_{(k)}$ is called the $k$-th $(k=1,\cdots,n)$ order statistics~\cite{David:2004}. The \textit{probability density function} (PDF) of $X_{(k)}$ is given by:
\[
\phi_{(k)}(\hat{x}) = \frac{n!}{(k-1)!(n-k)!}\Phi^{k-1}(\hat{x}) \phi(\hat{x}) [1-\Phi(\hat{x})]^{n-k}
\]
By condition on $n=2$, yielding:
\begin{eqnarray}
	\phi_{(1)} = 2\phi(\hat{x}) [1-\Phi(\hat{x})] \label{eq:order1},~~~
	\phi_{(2)} = 2\phi(\hat{x}) \Phi(\hat{x}) \label{eq:order2}
\end{eqnarray}
\par
\begin{figure}[h]
	\centering
	\includegraphics[scale=0.2]{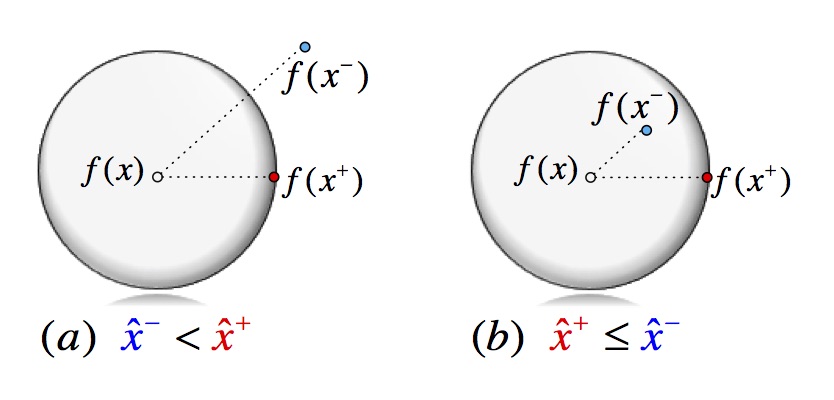}
	\caption{Two possible cases for the relative positions of anchor, positive, and negative triples.}
	\label{Fig:openball}
\end{figure}

The order statistics are determined by the distribution $\phi$ and can be regarded as properties of $\phi$. Next, we will examine the relative position of positive and negative samples to gain deeper insights into the class conditional density of true negatives $\phi_{\textsc{Tn}}$. Considering a triple $(x, x^+, x^-)$, two possible cases exist: (a) $\hat{x}^- < \hat{x}^+$, and (b) $\hat{x}^+ \leq \hat{x}^-$. In terms of order statistics notation, $\hat{x}^-$ (marked in blue in Fig~\ref{Fig:openball}) represents a realization of $X_{(1)}$ for case (a), or $\hat{x}^-$ represents a realization of $X_{(2)}$ for case (b), respectively.

\par
The generation process of observation $\hat{x}$ from $\phi_{\textsc{Tn}}$ can be described as follows: Select case (a) with probability $\alpha$, and then generate an observation $\hat{x}$ from $\phi_{(1)}$; Or select case (b) with probability $1-\alpha$, and then generate an observation $\hat{x}$ from $\phi_{(2)}$. That is, $\phi_{\textsc{Tn}}$ is the component of $X_{(1)}$ and $X_{(2)}$ with a \textit{mixture coefficient} $\alpha$
\begin{eqnarray}
	\phi_{\textsc{Tn}}(\hat{x}) = \alpha \phi_{(1)}(\hat{x}) + (1-\alpha)\phi_{(2)}(\hat{x}) \label{eq:phitn}
\end{eqnarray}
Similarly, $\phi_{\textsc{Fn}}$ is the component of $X_{(2)}$ and $X_{(1)}$ with mixture coefficient $\alpha$:
\begin{eqnarray}
	\phi_{\textsc{Fn}}(\hat{x}) = \alpha \phi_{(2)}(\hat{x}) + (1-\alpha)\phi_{(1)}(\hat{x})\label{eq:phifn}
\end{eqnarray}
Note that the way of taking $\hat{x}^-$ as a realization of $X_{(1)}$ for case (a) omits the situation of $\hat x^- = \hat x^+$. The probability measure of $\hat x^-$ for such case is 0 as $\phi$ is continuous density function dominated by Lebesgue measure. 

\begin{figure}[!]
	\centering
	\includegraphics[width=0.5\textwidth]{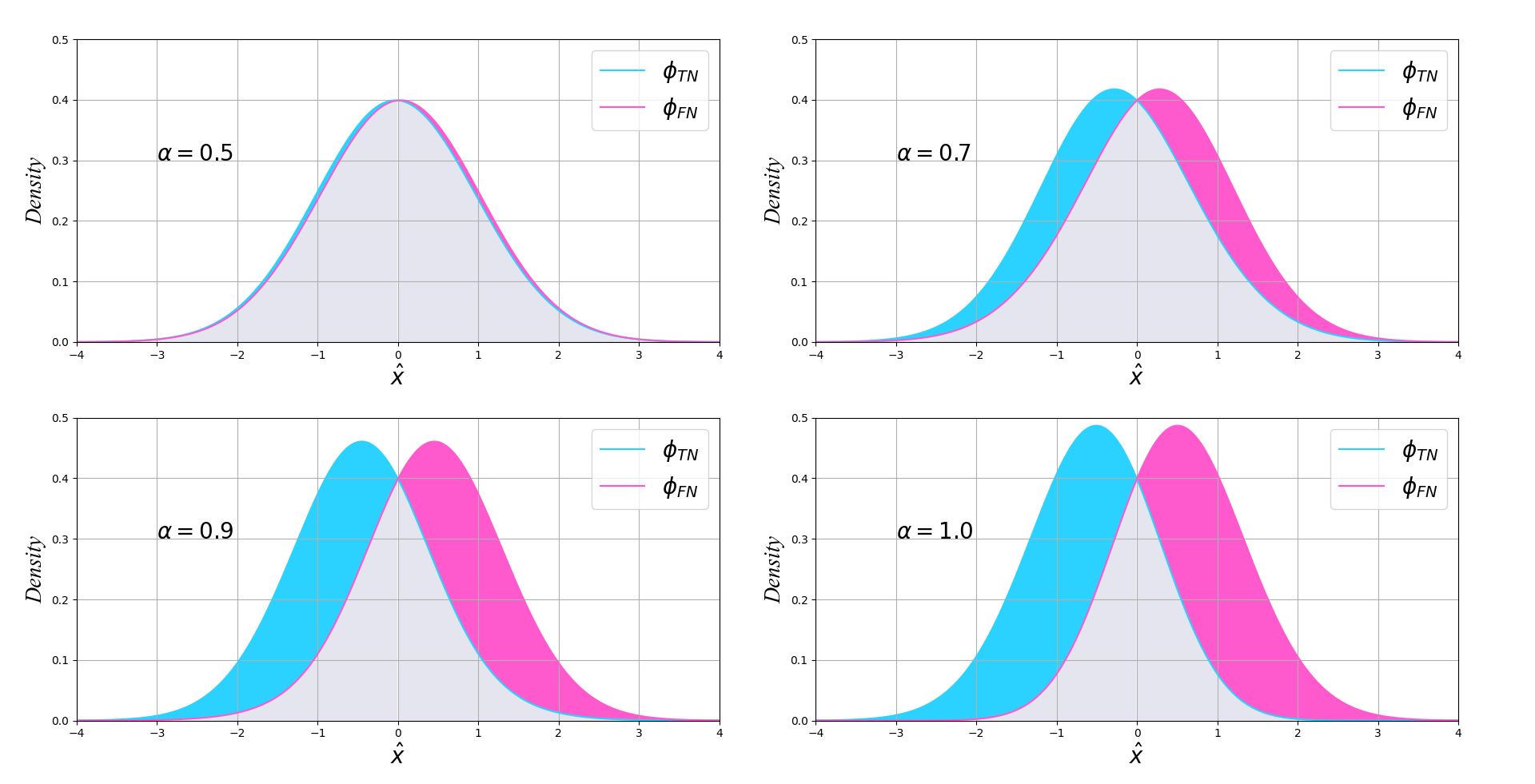}
	\caption{$\phi_{\textsc{Tn}}(\hat{x})$ and $\phi_{\textsc{Fn}}(\hat{x})$ with different $\alpha$ settings.}
	\label{Fig:theorydist}
\end{figure}

\begin{proposition}[Class Conditional Density] \label{proposition:1}
	If $\phi(\hat x)$ is continuous density function that satisfy $\phi(\hat x) \geq 0 $ and $\int_{-\infty}^{+\infty}\phi(\hat x) d\hat x =1 $, then  $\phi_{\textsc{Tn}}(\hat{x})$ and $\phi_{\textsc{Fn}}(\hat{x})$ are probability density functions that satisfy $\phi_{\textsc{Tn}}(\hat{x})\geq 0$, $\phi_{\textsc{Fn}}(\hat{x})\geq 0$, and $\int_{-\infty}^{+\infty}\phi_{\textsc{Tn}}(\hat{x})d\hat x =1 $, $\int_{-\infty}^{+\infty}\phi_{\textsc{Fn}}(\hat{x})d\hat x =1 $.
	\begin{proof}
		See Appendix~\ref{Proof:ccd}.
	\end{proof}
\end{proposition}
By setting $\phi(\hat{x})$ as $\mathcal{N}(0,1)$ in Eq.~\eqref{eq:order1} and Eq.~\eqref{eq:order2}, we can get a quick snapshot of how $\alpha$ affects $\phi_{\textsc{Tn}}(\hat{x})$ and $\phi_{\textsc{Fn}}(\hat{x})$ in Eq.~\eqref{eq:phitn} and Eq.~\eqref{eq:phifn} as illustrated in Fig~\ref{Fig:theorydist}. The larger value of $\alpha$ results in higher discrimination of $\phi_{\textsc{Tn}}(\hat{x})$ from $\phi_{\textsc{Fn}}(\hat{x})$, since a better encoder encodes dissimilar data points with different class labels more orthogonal~\cite{Chuang:2020:NIPS}.

There may need a further understanding and clarification of the mixture coefficient $\alpha$ by reviewing Fig.~\ref{Fig:openball}. Intuitively, $\alpha$ is the probability that the model scores positive sample higher than that of negative sample. For a worst encoder $f$ that randomly guesses, $\alpha=0.5$; While for a perfect encoder $\alpha=1$. Therefor, the reasonable value of $\alpha \in [0.5, 1]$. In fact, $\alpha$ reflects the encoder's capability of scoring a positive sample higher than that of a negative sample, which admits the empirical macro-AUC metric over all anchors $x$ in the training data set $\mathcal{D}$:
\begin{eqnarray}
	\small
	\alpha &=& \int\limits_{x \in \mathcal{X}}  \int\limits_{0}^{+\infty} \int\limits_{0}^{+\infty} \mathbb I (\hat{x}^+ \ge \hat{x}^-)  p(\hat{x}^+ , \hat{x}^-) p(x) d\hat{x}^+  d\hat{x}^- dx\nonumber\\
	&\simeq& \frac{1}{|\mathcal{D}|} \frac{1}{|\mathcal{D}^+| |\mathcal{D}^-|} \sum_{\mathcal{D}^+}\sum_{\mathcal{D}^-}  \mathbb I (\hat{x}^+ \ge \hat{x}^-) \nonumber \\
	&=& \frac{1}{|\mathcal{D}|} AUC
\end{eqnarray}

The class-conditional distribution $\phi_{\textsc{Tn}}(\hat{x})$ of true negative scores is the ideal target sampling population. The loss function computed from this subset of samples is the supervised loss. The first term of $\phi_{\textsc{Tn}}(\hat{x})$, $\alpha  \phi_{(1)}(\hat{x})$, is referred to as the \textit{easy negative component}, as illustrated by examples like $x_5^\prime$ (cat) in Fig~\ref{Fig:illustrative}. It corresponds to the distribution described in Fig~\ref{Fig:openball}(a), where true negative examples are pushed farther away from the anchor point and are correctly classified. The second term, $(1-\alpha)\phi_{(2)}(\hat{x})$, is referred to as the \textit{hard negative component}, as exemplified by $x_1^\prime$ (wolf) in Fig~\ref{Fig:illustrative}. It corresponds to the scenario described in Fig~\ref{Fig:openball}(b), where true negative samples are closer to the anchor point and are misclassified. To increase the sampling of the hard negative component, we introduce a hard negative mining parameter $\beta \in [0.5, 1]$. This parameter allows for the subdivision of the true negative population $\phi_{\textsc{Tn}}(\hat{x})$, where a proportion $\beta$ of samples is sampled from the \textit{hard negative component}, while a proportion $1-\beta$ is sampled from the \textit{easy negative component}, thereby creating a new sampling target. A $\beta$ value of 0.5 means that the two components are sampled in equal proportions, without any mechanism for hard negative mining.
\begin{eqnarray}
	\phi_{\textsc{Thn}}(\hat{x}) 
	&=& (1-\beta)\alpha  \phi_{(1)}(\hat{x}) +\beta (1-\alpha)  \phi_{(2)}(\hat{x}) \label{eq:tnhard} 
\end{eqnarray}
Eq~\eqref{eq:tnhard} can be interpreted as the distribution of hard negative samples with a hardness level of $\beta$. A higher value of $\beta$ (close to 1) results in a higher proportion of hard negative component in target sampling population.
\subsubsection{Actual Sampling Population}
With the class conditional density of positive samples and negative samples, we can obtain the density of unlabeled data $\phi_{\textsc{Un}}$ by applying total probability formula to relate marginal density to conditional densities
\begin{eqnarray}\label{eq:unlabel}
	\phi_{\textsc{Un}} &=& \tau^-\phi_{\textsc{Tn}} +\tau^+\phi_{\textsc{Fn}}  
\end{eqnarray} 
Note that $\phi_{\textsc{Un}}$ is the actual sampling population. The samples from $\phi_{\textsc{Un}}$ are directly used as negative sample to calculate contrastive loss. Although we do not know its explicit expression, we can compute its corresponding empirical cumulative distribution function using N observed unlabeled samples:
\begin{equation}\label{eq:puedf}
	\hat \Phi_{\textsc{Un}} (\hat{x}) = \frac{1}{n} \sum_{i=1}^{n}\mathbb{I}_{|X_i \leq \hat{x}|},
\end{equation}
As an example, let's consider that we have observed similarity scores of five unlabeled samples: $\{6, 4, 3, 7, 5\}$. In this case, $\hat \Phi_{\textsc{Un}} (6) = 4/5$, which means that the proportion of samples with scores less than or equal to 6 is 4/5. Above empirical cumulative distribution function (eCDF) $\hat \Phi_{\textsc{Un}} (\hat{x})$ uniformly converges to common cumulative distribution function (CDF) $\Phi_{\textsc{Un}}(\hat{x}) = \int_{-\infty}^{\hat{x}} \phi_{\textsc{Un}}(t)dt $ as stated by Glivenko theorem~\cite{glivenko:1933}. By leveraging the powerful tool of eCDF, we can characterize the distribution of neural network outputs and avoid making parameterized assumptions. To avoid confusion, we have listed the distributions involved in Table~\ref{table:density}.
\begin{table}[!]
	\centering
	\caption{Overview of densities.}\label{table:density}
	\resizebox{0.5\textwidth}{!}{
		\begin{tabular}{llc}
			\toprule[1.2pt]	
			Densities&Meaning& Parametric Assumption \\ \hline		
			$\phi(\hat{x})$ & Anchor specific distribution of scores. & No \\
			$\phi_{(1)}(\hat{x})$ &Order statistics $X_{(1)}$& No\\
			$\phi_{(2)}(\hat{x})$ &Order statistics $X_{(2)}$& No \\\hline
			$\phi_{\textsc{Tn}}(\hat{x})$&Class conditional distribution of TN scores & No \\
			$\phi_{\textsc{Fn}}(\hat{x})$&Class conditional distribution of FN scores&  No \\
			$\phi_\textsc{Thn}(\hat{x})$& Class conditional distribution of hard TN scores.& No  \\\hline
			$\phi_{\textsc{Un}}(\hat{x})$& Distribution of observed unlabeled scores.& No  \\
			\hline
			\bottomrule[1.2pt]
			
			~		\end{tabular}
	}
\end{table}

\subsubsection{Monte Carlo Importance Sampling}
Now we have N i.i.d unlabeled samples $\{\hat{x}_i\}_{i=1}^N \sim \phi_{\textsc{Un}} $, and desired target sampling distribution $\phi_\textsc{Thn}$, we can calculate importance weight for re-weighting unlabeled sample $\hat x_i$ using classic Monte-Carlo importance sampling~\cite{Hesterberg:1988,Bengio:2008:TNN}:
\begin{small}
	\begin{eqnarray}
		&&\omega_i(\hat x_i;\alpha, \beta)\nonumber\\
		&=& \frac{\phi_\textsc{Thn}(\hat{x};\alpha, \beta)/Z}{\phi_{\textsc{Un}}(\hat{x})} \label{eq:ome} \\
		&=& \frac{1}{Z} \frac{(1-\beta)\alpha + (\beta-\alpha)\Phi(\hat{x})}{\alpha\tau^- + (1- \alpha)\tau^+  + (1-2\alpha)\Phi(\hat{x})(\tau^- -\tau^+)}  \label{eq:ome1}
	\end{eqnarray}
\end{small}
where  $\phi_\textsc{Thn}$ is the target sampling population, $\phi_{\textsc{Un}}(\hat{x})$ is the actual sampling population.  $Z$ is the normalization constant of the target sampling distribution $\phi_\textsc{Thn}$, which can be accurately computed through marginal integration as $Z = \int_{ - \infty } ^{\infty} \phi_{\textsc{Thn}}(\hat{x})  d \hat{x} = (1-\beta)\alpha + \beta (1-\alpha)$. 

$\omega_i(\hat x_i;\alpha, \beta)$ is a function of C.D.F $\Phi (\hat{x})$, while $\phi{(\hat{x})}$ is eliminated due to the fractional form of $\omega$. We may not be able to directly compute $\Phi(\hat{x})$, but we can calculate $\Phi_{\textsc{un}}(\hat{x})$. 
By integrating both sides of the Eq~\eqref{eq:unlabel} we can establish the transformation of two cumulative distribution functions $\Phi(\hat{x})$ and $\Phi_{\textsc{Un}}(\hat{x})$ as follows:
\begin{eqnarray}\label{eq:cdftrans}
	\Phi(\hat{x}) = \frac{-b+\sqrt{b^2+4a\Phi_{\textsc{Un}}(\hat{x})}}{2a}
\end{eqnarray}
where $ a=  (1-2\alpha)(\tau^- -\tau^+)$ and $b = 2[\alpha\tau^- + (1- \alpha)\tau^+]$. The details are presented in Appendix~\ref{sec:trans}. 

So far, we have completed the computation of the importance weight. The importance weight $\omega(\hat{x}, \alpha, \beta)$ assigns customized weights to the N unlabeled samples to correct the bias caused by sampling from the unlabeled data. It involves two parameters: the false negative debiasing parameter $\alpha$, which represents the AUC of the encoder and can be empirically estimated from a small portion of the validation set, and the hard negative mining parameter $\beta$, which is a task-specific hyperparameter. The computation of the importance weight involves two types of information: the class prior information $\tau$ and the eCDF of the unlabeled samples $\hat \Phi_{\textsc{Un}} (\hat{x})$. The eCDF represents the sample information from a Bayesian perspective and reflects the model's discriminative results. The higher the probability that the model classifies an unlabeled sample as a positive sample, the higher its similarity score will be, and the closer $\hat \Phi_{\textsc{Un}} (\hat{x})$ will be to 1. Therefore, we can interpret $\hat \Phi_{\textsc{Un}} (\hat{x})$ as the likelihood of an unlabeled sample being a false negative. Thus, the computation of the weight incorporates the posterior information of the unlabeled samples being true negatives. In Lemma~\ref{lemma:1}, we will analyze its connection with posterior probability. So we refer to as the \textbf{B}ayesian \textbf{S}elf-\textbf{S}upervised \textbf{C}ontrastive \textbf{L}oss.

\begin{algorithm}[!]
	\caption{The proposed method.}\label{alg:bcl}
	\KwIn{$\alpha$, $\beta$,  one score of positive sample $e^{f(x)^Tf(x^+)}$, $N$ scores of unlabeled samples $\{e^{f(x)^Tf(x^\prime_i)}\}_{i=1}^N$.}
	\KwOut{BCL loss.}
	\For{$i=1,...N$}{
		~~Calculate eCDF $\hat \Phi_{\textsc{Un}} (\hat{x}_i)$ by Eq.~\eqref{eq:puedf}; \\
		Calculate $\Phi(\hat{x}_i) $ by Eq.~\eqref{eq:cdftrans}; \\
		Calculate $\omega_i(\hat{x}_i;\alpha, \beta)$ by Eq~\eqref{eq:ome1};
	}
	Calculate $\mathcal{L}_\textsc{~Bcl}$ by Eq.~\eqref{eq:debias};
	
	\KwResult{$\mathcal{L}_\textsc{~Bcl}$}
\end{algorithm}
\textbf{Complexity}: Compared with the InfoNCE loss, the additional computation complexity comes from first calculating the eCDF $\hat \Phi_{\textsc{Un}} (\hat{x})$ from unlabeled data by Eq.~\eqref{eq:puedf} in the order of $\mathcal{O}(N)$, where $N$ is the number of negative samples. The additional computation complexity $\mathcal{O}(N)$ can be neglected as it is far smaller than encoding a sample. We compared the actual running time of BCL with InfoNCE, and found that they are almost identical.\footnote{The source codes has been  released at \url{https://anonymous.4open.science/r/BCL-2A1B/README.md}}

\subsubsection{Theoretical Analysis} 
We investigate the small-sample properties and large-sample properties of BCL under three lenient conditions as depicted in Fig.~\ref{fig:condition} in Appendix: (i) Let the similarity score of unlabeled samples $\hat{x}$ be independently and identically distributed according to an unknown distribution $\phi$. (ii) Among them, the proportion of positive samples $\hat{x}^+$ is $\tau^+$, and (iii) $\mathbb{P}(\hat{x}^- < \hat{x}^+) = \alpha$, that is, the AUC of current encoder is $\alpha$. 

\begin{lemma}[Posterior Probability Estimation]\label{lemma:1} 
	Under the aforementioned conditions, with $\beta = 0.5$, the  weight $\omega_i$ computed using Eq.~\eqref{eq:ome1} and the posterior probability of the sample being a true negative $\mathbb P(\textsc{Tn}|\hat{x}_i)$ satisfy $\omega_i = \mathbb{P}(\textsc{Tn}|\hat{x}_i)/{\tau^-}$.
	\begin{proof}
		See Appendix~\ref{Proof:poster}
	\end{proof}
\end{lemma}
Lemma~\ref{lemma:1} implies that as the posterior probability of a sample being a true negative increases, a larger weight $\omega_i$ will be calculated, thereby pushing away hard negatives in accordance with the principle of hard negative mining. Conversely, if the posterior probability of a sample being a true negative decreases, a smaller weight will be computed to prevent positive samples from being pushed apart, following the principle of false negative debiasing. 
It addresses the two critical tasks in a flexible and principled way to 
align the semantic of negative samples across pre-training and downstream tasks, and sets our work apart from existing methods based on importance weighting techniques~\cite{Chuang:2020:NIPS,Robinson:2021:ICLR,wu2023understanding}.

\begin{lemma}[Asymptotic Unbiased Estimation] \label{lemma:2}
	Under the aforementioned conditions, with $\beta = 0.5$, for any encoder $f$, as $N\rightarrow \infty$, we have $\mathcal{L}_\textsc{~Bcl}\rightarrow\mathcal{L}_\textsc{~Sup}$.
	\begin{proof}
		See Appendix~\ref{Proof:unbias}
	\end{proof}
\end{lemma}
The core idea of the proof is to leverage the properties of importance sampling by using the weighted similarity scores of unlabeled samples (from the actual sampling population), denoted as $\sum_{i=1}^N \omega_i \cdot \hat{x}_i$, to approximate the expected value of similarity scores of true negative samples (from the target sampling population). Lemma~\ref{lemma:2} demonstrates that as $N$ approaches infinity, the limit form of the BCL loss, denoted as $\mathcal{L}^{~\text{N}}_\textsc{~Bcl}$, is consistent with supervised contrastive loss defined by Eq~\eqref{Eq:SupLoss}. Obtaining an estimate that is consistent with the supervised loss is an important approach to improving the generalization performance of weakly supervised or unsupervised learning, as the loss function is a function of model parameters $\Theta$ and training samples $\mathbf{x}$, denoted as $\mathcal{L}(\mathbf{x},\Theta)$. If, for a sufficient large number of training samples $\mathbf{x}$, the unsupervised loss $\mathcal{L}(\mathbf{x},\Theta)$ is equal to the supervised loss $\mathcal{L}_\textsc{Sup}(\mathbf{x},\Theta)$ almost everywhere for all possible $\Theta$, then the unsupervised loss $\mathcal{L}(\mathbf{x},\Theta)$ and the supervised loss $\mathcal{L}_\textsc{Sup}(\mathbf{x},\Theta)$ share the same extreme points $\Theta$. As a result, learning in the unsupervised case can lead to model parameters that are close to those learned under supervised learning, thereby achieving similar generalization performance as supervised learning. However, in practice, only a finite value of N can be selected. Lemma~\ref{lemma:3} bounds the estimation error due to finite $N$ as decreasing in the order of $\mathcal{O}(N^{-1/2})$.

\begin{lemma}[Estimation Error Bound] \label{lemma:3}
	Under the aforementioned conditions, with $\beta = 0.5$, for any encoder $f$ and $N> 0$,
	$|\hat{\mathcal{L}}_\textsc{~Bcl}-\mathcal{L}^{~\text{N}}_\textsc{~Bcl}|\leq \tau^- \sqrt{\frac{2\pi}{N}}$.
	\begin{proof}
		See Appendix~\ref{Proof:bound}
	\end{proof}
\end{lemma}

\section{Experiments}
\subsection{Numerical Experiment}
BCL uses the weighted similarity scores of unlabeled samples to approximate the expected value of similarity scores of true negative samples. In the numerical experiments section, while evaluating the quality of the BCL estimator, we also gain a better understanding of the underlying generation mechanism of observed scores of unlabeled samples. In the case of supervised contrastive learning, the mean value of observations for true negative samples calculated by
\begin{equation}\label{eq:true}
	\theta_{\textsc{Sup}} = {\sum_{i=1}^N \mathbb{I}(x_i) \cdot \hat{x}_i  }/{ \sum_{i=1}^N \mathbb I(x_i) },
\end{equation}
where $\mathbb I(x_i)$ is the indicator function, $\mathbb I(x_i)=1$ if $x_i \in \textsc{Tn}$; Otherwise, $\mathbb I(x_i)=0$. We compare $\hat{\theta}_\textsc{Bcl}$ with $\theta_{\textsc{Sup}}$ to evaluate the quality of BCL estimator, since the expectation is replaced by empirical estimates in practice~\cite{Oord:2018:arxiv,Chuang:2020:NIPS,Chen:2020:ICML}. Additionally, two other commonly used estimates are considered: $\hat{\theta}_{\textsc{Biased}}$ by~\cite{Oord:2018:arxiv} and $\theta_{\textsc{Dcl}}$ by~\cite{Chuang:2020:NIPS}.
\begin{small}
\begin{eqnarray}
	\hat{\theta}_\textsc{Biased} =\frac{1}{N} \sum_{i=1}^{N} \hat{x}_i \label{Eq:BiasedEstimator},
	\hat{\theta}_\textsc{Dcl} =  \frac{1}{N\tau^-}  (\sum_{i=1}^{N} \hat{x}_i - N\tau^+ \cdot \frac{\sum_{j=1}^{K} \hat{x}_j^+}{K} ).\label{Eq:DCLEstimator} \nonumber
\end{eqnarray}
\end{small}


\begin{figure}[!]
	\centering
	\includegraphics[width=0.5\textwidth]{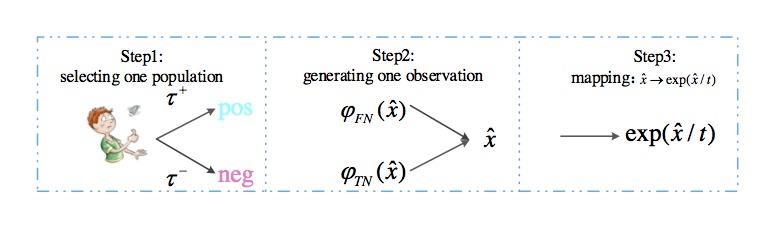}
	\caption{Generation process of observation $\exp(\hat{x}/t)$.}
	\label{fig:generateN}
\end{figure}

The generation process of similarity scores for unlabeled samples can be divided into three steps, as illustrated in Fig~\ref{fig:generateN}: \textbf{(i)} Select a class according to the class prior $\tau^+$, indicating that the observation comes from $\textsc{Fn}$ with probability $\tau^+$, or comes from $\textsc{Tn}$ with probability $\tau^-$. \textbf{(ii)} Sample an observation $\hat{x}$ from the class conditional density $\phi_{\textsc{Fn}}$ (or $\phi_{\textsc{Tn}}$). \textbf{(iii)} Map $\hat{x}$ to $\exp(\hat{x}/t)$ as the final observation.

\par
The complete stochastic process depiction of observations is presented in Algorithm~\ref{Alg:numer} of Appendix~\ref{Sec:numer}. Note that in step \textbf{(ii)}, sampling the observations from $\phi_{\textsc{Fn}}$ (or $\phi_{\textsc{Tn}}$) is achieved by using the accept-reject sampling~\cite{casella2004generalized} technique (see  Algorithm~\ref{Alg:AccRetTN} and ~\ref{Alg:AccRetFN} of Appendix~\ref{Sec:numer} for implementation details). Note that in \textbf{(iii)}, $\hat{x}\rightarrow \exp(\hat{x}/t)$ is a strictly monotonic transformation, making the empirical distribution function remains the same: $\hat \Phi_\textsc{Un}(\hat{x}) = \hat \Phi_\textsc{Un}(\exp(\hat{x}/t))$.
\par

\begin{figure}[t]
	\centering
	\includegraphics[width=0.5\textwidth]{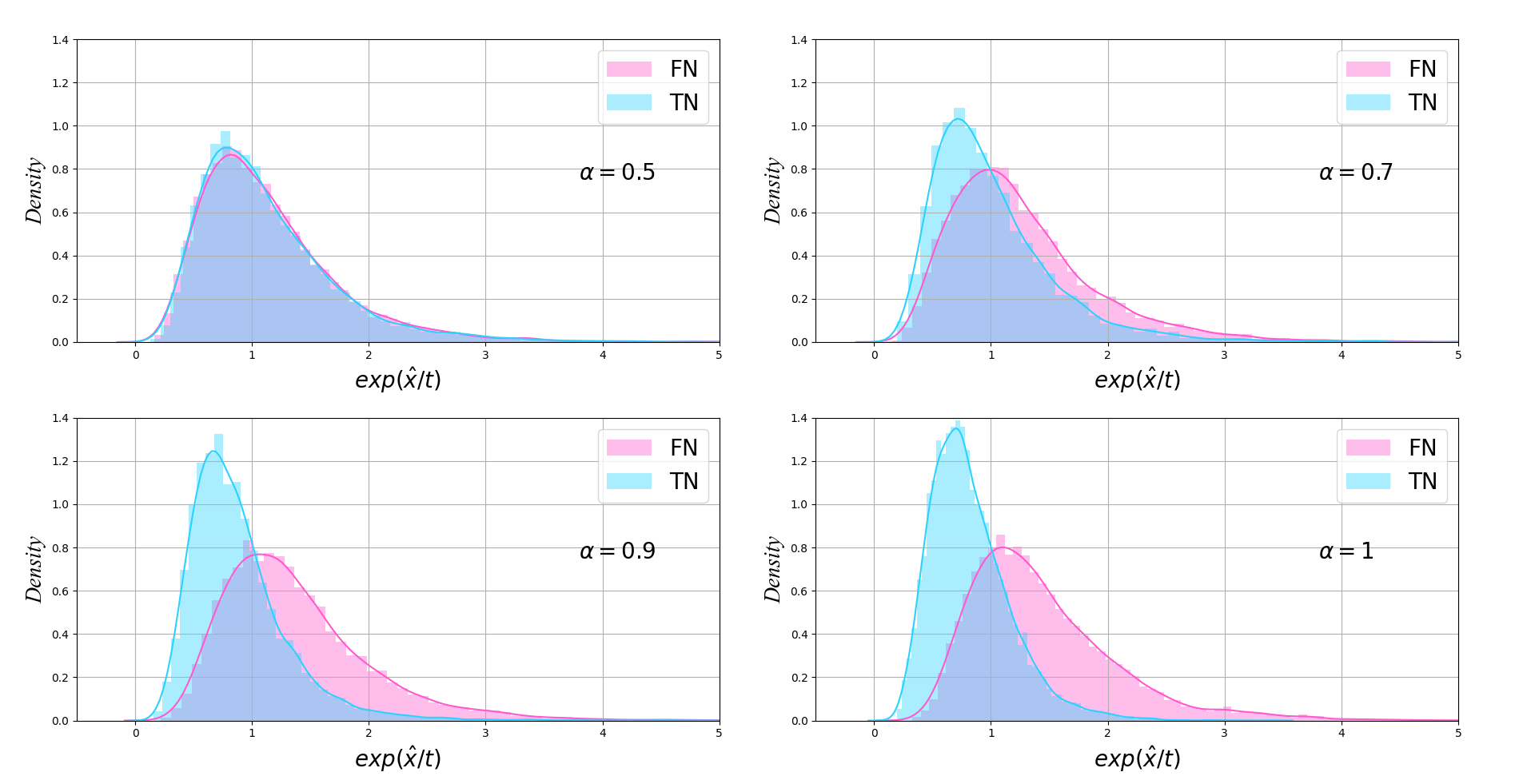}
	\caption{Empirical distribution of $\exp(\hat{x}/t)$  with different $\alpha$.}
	\label{Fig:empiricaldist}
\end{figure}
We initialize the distribution $\phi$ as $\mathcal{N}(0,1)$, and then repeat  Fig~.\ref{Fig:illustrative} $N$ times, then we can generate $N$  similarity scores for labeled samples. Fig.~\ref{Fig:empiricaldist} displays the empirical distribution of $\exp(\hat{x}/t)$ according to the above generative process, which corresponds to the theoretical distribution depicted in Fig~.\ref{Fig:theorydist}, where we set $t=2$, $M=1$ and $N=20000$. We note that the empirical distributions in Fig~\ref{Fig:empiricaldist} exhibits similar structures to those by using real-world datasets as reported in~\cite{Robinson:2021:ICLR, Xia:2022:ICML}, indicating the effectiveness of the proposed method for simulating $\hat{x}$.

We then initialize the distribution $\phi$ as $U(a,b)$, where $-1/t^2\leq a \leq b \leq 1/t^2$ is randomly selected for each anchor. This choice allows for convenient control over the observations within its theoretical minimum and maximum interval of $[-1/t^2,1/t^2]$\cite{Chuang:2020:NIPS}. In our case, we initially set $a=-0.5$ and $b=0.5$. Additionally, we use $\gamma \in [0,1]$ to control the maximum sliding amplitude of $U(a,b)$ from the original interval for each anchor.

We evaluate the quality of estimators in terms of \textit{mean squared error} (MSE). For $M$ anchors, we calculate $\textsc{Mse}~\hat{\theta}_{\textsc{Bcl}} = \frac{1}{M}\sum_{m=1}^M (\hat{\theta}_{\textsc{Bcl}, m} -\theta_{\textsc{Sup}, m})^2$.  Fig.~\ref{Fig:MSE} in Appendix~\ref{Sec:numer} compares the MSE of different estimators against different parameters.
It can be observed that the estimator $\hat{\theta}_{\textsc{Bcl}}$ is superior than the other two estimators in terms of lower MSE with different parameter settings. We refer to Appendix~\ref{Sec:numer} for detailed settings.

\subsection{Image Classification}
Since the weight $\omega$ acts on the gradients of negative samples, our comparative method is restricted to end-to-end parameter update methods, with SimCLR serving as a representative approach. The baselines used in our experiments include: \textbf{SimCLR}\cite{Chen:2020:ICML}: It provides the experiment framework for all competitors, which comprises of a stochastic data augmentation module, the ResNet-50 as an encoder, neural network projection head. It uses the contrastive loss $\mathcal{L}_\textsc{Biased}$. \textbf{DCL}\cite{Chuang:2020:NIPS}: It mainly considers the false negative debiasing task and uses the $\hat{\theta}_\textsc{Dcl}$ estimator in Eq.\eqref{Eq:DCLEstimator} to replace the second term in the denominator of $\mathcal{L}_\textsc{Biased}$. \textbf{HCL}\cite{Robinson:2021:ICLR}: Following the DCL debiasing framework, it also takes into consideration of hard negative mining by up-weighting each randomly selected sample as $
\omega_i^\textsc{Hcl} = \frac{\hat{x}i^\beta}{\frac{1}{N} \sum{j=1}^{N} \hat{x}_j^\beta}
$. Note that DCL is a particular case of HCL with $\beta=0$. Our experimental setup is in line with the comparative methods, ensuring a fair comparison across CIFAR10, STL10, CIFAR100, and TinyImageNet datasets for image classification tasks.

\textbf{Binary image classification}: We firstly investigate binary classification task that with higher rate of false negative examples ($\tau^+=0.5$). For each datasets, we randomly select two classes along with their corresponding samples. Then, we train the ResNet50 encoder using four different learning algorithms. The top-1 accuracy is shown in the following table.
\begin{table}[h!]
	\centering
	\caption{Top-1 accuracy for binary image classification.}\label{Exp:result2}
	\resizebox{0.45\textwidth}{!}{
		\begin{tabular}{lcclccc}
			\toprule[1.2pt]
			\textbf{Datasets} &Classes&	N&	SimCLR&	DCL &	HCL&	\textbf{BCL}\\\hline
			CIFAR10&2&	510	&99.1	&98.1&	98.4&	\textbf{99.3}($\pm 0.22$)\\
			STL10	&2&510&	91.7&	91.2&	91.1&	\textbf{92.7}($\pm 0.34$)\\
			CIFAR100&2&	510	&97.5&	86.9&	96.5&	\textbf{98.8}($\pm 0.28$)\\
			tinyImageNet&2&	510&	97.0&	96.0&	94.0&\textbf{99.0}($\pm 0.26$)\\
			\cline{1-7}
			\bottomrule[1.2pt]
			
		\end{tabular}
	}
\end{table}

\textbf{Multi-class image classification}: Next, we consider multi-classification tasks with a lower positive class prior ($\tau^+=1/C$). We train the ResNet50 encoder using all the available samples. The experimental results are shown below.
\begin{table}[h!]
	\centering
	\caption{Top-1 accuracy for multi-class image classification.}\label{Exp:result1}
	\resizebox{0.45\textwidth}{!}{
		\begin{tabular}{lcclccc}
			\toprule[1.2pt]
			\textbf{Datasets} &Classes&	N&	SimCLR&	DCL &	HCL&	\textbf{BCL}\\\hline
			CIFAR10	&10&510&	91.1&	92.1&	91.9&	\textbf{92.5}($\pm 0.15$)\\
			STL10&10&	510	&80.2	&84.3&	87.4&	\textbf{87.7}($\pm 0.12$)\\
			CIFAR100&100&	510	&66.4&	67.7&	69.5&	\textbf{69.7}($\pm 0.10$)\\
			tinyImageNet&200&	510&	53.4&	53.7&	57.0&\textbf{57.3}($\pm 0.09$)\\
			\cline{1-7}
			\bottomrule[1.2pt]
			
		\end{tabular}
	}
\end{table}

For N = 510 (batch size=256) negative examples per data point, we observed absolute improvements of 0.2\%, 1.0\%,1.3\%, and 2.0\% over SimCLR in the binary classification task. Furthermore, we observed absolute improvements of 1.4\%, 7.7\%, 3.3\%, and 2.9\% over SimCLR in the multi-class image classification task, respectively. It is noteworthy that in the binary classification task, SimCLR generally achieves the second-best results, while DCL and HCL perform relatively poorly. We attribute this result to the inflexibility of their debiasing mechanisms. When $\tau$ is too large, the denominator in DCL's debiasing mechanism subtracts a significantly larger value. In the multi-class classification task, the second-best results are typically achieved by HCL, which incorporates hard negative mining mechanism. With a larger number of classes, indicating a smaller class prior probability, there should be a stronger emphasis on the task of hard negative mining.

\textbf{Hyperparameter analysis}: Table~\ref{Exp:N} presents the classification accuracy of BCL for different  $N$. Similar to SimCLR, larger values of $N$ lead improved performance. With the same batch size, BCL significantly outperforms SimCLR while maintaining a identical runtime.

Table~\ref{Exp:beta} presents the impact of different hardness level $\beta$ on the performance of BCL. On the STL10 dataset, the top-1 ACC continues to increase with increasing $\beta$. However on the CIFAR10 dataset, the top-1 classification accuracy gradually increases as beta increases, reaching its optimal value at 0.9 before decreasing. This result indicates that a higher hardness level may not always lead to better performance. This is further supported by the fact that HCL, which includes a hard negative mining mechanism, performs worse than DCL (DCL is a particular case of HCL without hard negative mining mechanism) on the CIFAR10 dataset.

Regarding the setting of $\beta$, we can draw upon empirical findings from the field of noisy label learning. It has been observed that deep neural networks tend to first fit clean samples before fitting noisy samples\cite{Arpit:2017:ICML,zhang2021understanding, Han:2018:NIPS}. In this context, false-negative examples can be considered as noise present in the negatives. Larger values of $\beta$ lead to higher weights assigned to the hard samples, resulting in larger loss values and subsequently larger gradients for those samples. From the perspective of noise rate ($\tau^+$), we recommend to set $\beta=1-1/C$. Additionally, another factor that needs to be taken into account is the level of difficulty in fitting the dataset itself. The STL-10 dataset poses relatively more challenges compared to the CIFAR-10 dataset, as indicated by its much lower ACC1. Therefore, larger values of $\beta$ for STL10 can effectively encourage the neural network to fit the clean and hard samples before fitting noisy samples.

\begin{table}[!]
	\centering
	\caption{Impacts of negative sample size $N$.}\label{Exp:N}
	\resizebox{0.45\textwidth}{!}{
		\begin{tabular}{lclcccc}
			\toprule[1.2pt]
			\multirow{2}*{\textbf{Dataset}} &  \multirow{2}*{\textbf{Methods}} & \multicolumn{2}{c}{Negative Sample Size $N$} \\\cline{3-7}
			
			~ & ~ & N=30 & N=62& N=126& N=254&N=510 \\ \hline
			
			\multirow{2}*{\textbf{CIFAR10}} & SimCLR & 80.21 & 84.82   &87.58&89.87	&91.12(82s/epoch)\\
			~&BCL& 83.61 & 88.56   &90.83	&92.07	&92.58 (82s/epoch)\\
			\cline{1-7}
			
			\multirow{2}*{\textbf{STL10}} & SimCLR & 61.20 & 71.69   &74.36	&77.33	&80.20 (169s/epoch)\\
			~&BCL& 67.45 & 73.36   &80.23	&84.68	&87.65 (169s/epoch)\\
			\cline{1-7}
			\bottomrule[1.2pt]
			
		\end{tabular}
	}
\end{table}

\begin{table}[!]
	\centering
	\caption{Impact of hardness level $\beta$.}\label{Exp:beta}
	\resizebox{0.45\textwidth}{!}{
		\begin{tabular}{cclcccc}
			\toprule[1.2pt]
			
			\textbf{Dataset} & $\beta=0.5$ & $\beta=0.6$ & $\beta=0.7$ & $\beta=0.8$ & $\beta=0.9$ &$\beta=1.0$  \\ \hline
			\textbf{CIFAR10}&91.39 & 91.41 & 91.89   &92.02	&\textbf{92.58}	&92.12\\
			\textbf{STL10}&80.32& 81.79 & 83.58   &83.83	&84.85	&\textbf{86.51}\\
			\cline{1-7}
			\bottomrule[1.2pt]
			
		\end{tabular}
	}
\end{table}

\section{Conclusion}
This paper introduces a generalized view of contrastive loss called BCL loss for self-supervised contrastive learning. BCL provides the posterior probability estimation of unlabeled sample being true negative, and asymptotic unbiased estimation of supervised contrastive loss. Essentially, BCL modifies the gradient-based parameter updates through importance re-weighting. It is worth noting that asynchronous update methods, such as MOCO\cite{He:2020:CVPR} and BYOL\cite{BYOL:2020:NIPS}, employ momentum-based updates to scatter features, resulting in impressive performance. Our future research objectives include exploring the extension of gradient-based parameter updates to momentum-based methods and investigating improved momentum update techniques.


\textbf{Impact Statements}

This paper aimed at advancing the field of Machine Learning. It tackles the prominent challenge of semantic inconsistency in negative samples between the pretext task and downstream tasks, with the potential to enhance the performance of contrastive learning-based applications. Importantly, after thorough analysis, no discernible negative social impacts have been identified that require specific highlighting in this context.

\newpage
\nocite{langley00}
\bibliography{main}
\bibliographystyle{icml2024}

\newpage
\appendix
\onecolumn
The contents of the appendix are organized as follows:
\begin{itemize}
	\item Theoretical Analysis: Appendix~\ref{sec:proof}
	\subitem Sampling Bias Analysis: Appendix~\ref{sec:bias}
	\subitem Derivation of Equation~\eqref{eq:cdftrans}: Appendix~\ref{sec:trans}
	\subitem Proof of Proposition~\ref{proposition:1}: Appendix~\ref{Proof:ccd}
	\subitem Proof of Lemma~\ref{lemma:1}: Appendix~\ref{Proof:poster}
	\subitem Proof of Lemma~\ref{lemma:2}: Appendix~\ref{Proof:unbias}
	\subitem Proof of Lemma~\ref{lemma:3}: Appendix~\ref{Proof:bound}
	\item Numerical Experiments: Appendix~\ref{Sec:numer}
	\subitem Implementation Details of Accept-Reject Sampling: Appendix~\ref{appb:accrej}
	\subitem Parameter Settings: Appendix~\ref{appb:para}
	\item Image Classification: Appendix~\ref{sec:image}
	\subitem Data Statistics: Appendix~\ref{appc:data}
	\subitem Implementation Details: Appendix~\ref{appc:deta}
	\subitem Hyper Parameter Analysis: Appendix~\ref{appc:para} 
	\item Graph Embedding: Section~\ref{sec:graph}
	\subitem Experimental Setups: Appendix~\ref{appd:set}
	\subitem Data Statistics: Appendix~\ref{appd:data} 
	\subitem Experimental Results: Appendix~\ref{appd:res}
\end{itemize}
\section{Theoretical Analysis}\label{sec:proof}
\subsection{Sampling Bias Analysis}\label{sec:bias}
The analysis in this section aims to demonstrate that utilizing unlabeled samples as negative samples to optimize the contrastive loss in a self-supervised setting can lead to intractable optimization of mutual information.

Let $x^-\in \textsc{Tn}$ denote $x^-$ being a \textit{true negative} (TN) sample specific to $x$. Let $x^-\in \textsc{Fn}$ denote $x^-$ being a \textit{false negative} (FN) sample specific to $x$, i.e. $x^-$ and $x$ are with the same ground truth class label. Note that whether $x^-$ is a TN or FN is specific to a particular \textit{anchor} point $x$, and in what follows, we omit the \textit{specific to $x$} for brevity. It has been proven that for $\{x_i^- \in \textsc{Tn} \}_{i=1}^N$, optimizing the InfoNCE loss $\mathcal{L}_\textsc{Sup}$ will result in the learning model estimating and optimizing the \textit{density ratio} $\frac{p^+}{p^-}$~\cite{Oord:2018:arxiv,Ben:2019:ICML}. Denote $\hat{x}^+=e^{f(x)^Tf(x^+)}$. The CPC~\cite{Oord:2018:arxiv} proves that minimizing $\mathcal{L}_\textsc{Sup}$ leads to
\begin{equation}\label{Eq:LsupDensityRatio}
	\hat{x}^+ \propto p^+/p^-.
\end{equation}
As discussed by~\cite{Oord:2018:arxiv}, $p^+/p^-$ preserves the mutual information (MI) of future information and present signals, where MI maximization is a fundamental problem in science and engineering~\cite{Ben:2019:ICML,Belghazi:2018:ICML} .

\par
Now consider the InfoNCE loss $\mathcal{L}_\textsc{Biased}$, which can be regarded as the categorical cross-entropy of classifying one positive sample $x^+$ from unlabeled samples. For analysis purpose, we rewrite $x^+$ as $x_0$. Given $N+1$ unlabeled data points, the posterior probability of one data point $x_0$ being a positive sample can be derived by
\begin{eqnarray}
	P(x_0\in \textrm{pos}| \{x_i\}_{i=0} ^N) 
	&=& \frac{p^+(x_0) \prod_{i=1}^{N} p_d(x_i)}{\sum_{j=0}^{N} p^+(x_j) \prod_{i \neq j} p_d (x_i)} \nonumber \\
	&=& \frac{p^+(x_0)/p_d(x_0)}{p^+(x_0)/p_d(x_0) + \sum_{j=1}^{N} p^+(x_j)/p_d(x_j)}
\end{eqnarray}
To minimize $\mathcal{L}_\textsc{Biased}$, the optimal value for this posterior probability is 1, which is achieved in the limit of $p^+(x_0)/p_d(x_0) \rightarrow +\infty$ or $p^+(x_j)/p_d(x_j)\rightarrow 0$. Minimizing $\mathcal{L}_\textsc{Biased}$ leads to
\begin{eqnarray}\label{eq:propoto}
	\hat{x}^+ \propto p^+/p_d.
\end{eqnarray}
Note that this is different from Eq.~\eqref{Eq:LsupDensityRatio}, since $x_i^-$ may not be $\textsc{Tn}$ for lack of ground truth label.

\par
Denote $\hat{x}^+ = m\cdot p^+/p_d, ~m \ge 0$. We investigate the gap between optimizing $\hat{x}^+$ and the optimization objective $p^+/ p^-$. Inserting $p_d = p^-\tau^-+ p^+\tau^+ $ back to Eq.~\eqref{eq:propoto}, we obtain
\begin{eqnarray} \label{eq:gap}
	\hat{x}^+ = m\cdot \frac{p^+}{p^-\tau^-+ p^+\tau^+}.
\end{eqnarray}
Rearranging the above equation yields
\begin{equation}\label{eq:fractional}
	p^+/p^- = \frac{\hat{x}^+ \cdot \tau^-}{m-\hat{x}^+ \cdot \tau^+}.
\end{equation}
\par
\begin{figure}[h!]
	\centering
	\includegraphics[scale=0.5]{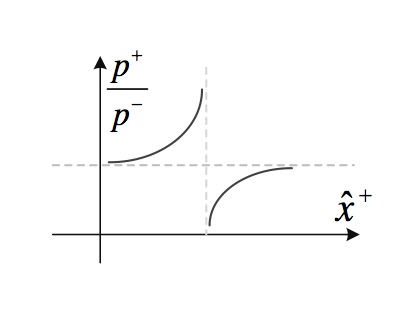}
	\caption{Illustration of $\mathcal{L}_\textsc{Biased}$ and mutual information optimization by Eq.~\eqref{eq:fractional}.}
	\label{Fig:gap}
\end{figure}
Fig.~\ref{Fig:gap} illustrates the approximate shape of Eq.~\eqref{eq:fractional} as a fractional function, which reveals the inconsistency between InfoNCE $\mathcal{L}_\textsc{Biased}$ loss optimization and MI optimization. That is, when optimizing InfoNCE loss, the increase of $\hat{x}^+$ does not lead to the monotonic increase of $p^+/p^-$. Indeed, the existence of \textit{jump discontinuity} indicates that the optimization of $\mathcal{L}_\textsc{Biased}$ does not necessarily lead to the tractable MI optimization. The reason for the intractable MI optimization is from the fact that not all $\{x_i^-\}_{i=1} ^N$ are $\textsc{Tn}$ samples, as they are randomly drawn from the data distribution $p_d$. This leads to the inclusion of $p^+$ in the denominator of Eq.~\eqref{eq:gap} when decomposing the data distribution $p_d$. Fig.~\ref{Fig:illustrative} in Appendix provides an intuitive explanation. The sampled data points actually contain one $\textsc{Fn}$ sample. Such a $\textsc{Fn}$ sample should be pulled closer to the anchor $x$. However, as it is mistakenly treated as a negative sample, during model training it will be pushed further apart from the anchor, which breaks the semantic structure of embeddings~\cite{Feng:2021:CVPR}.

\subsection{Derivation of Equation~\eqref{eq:cdftrans}}\label{sec:trans}
\begin{figure*}[h!]
	\centering
	\includegraphics[width=0.6\textwidth]{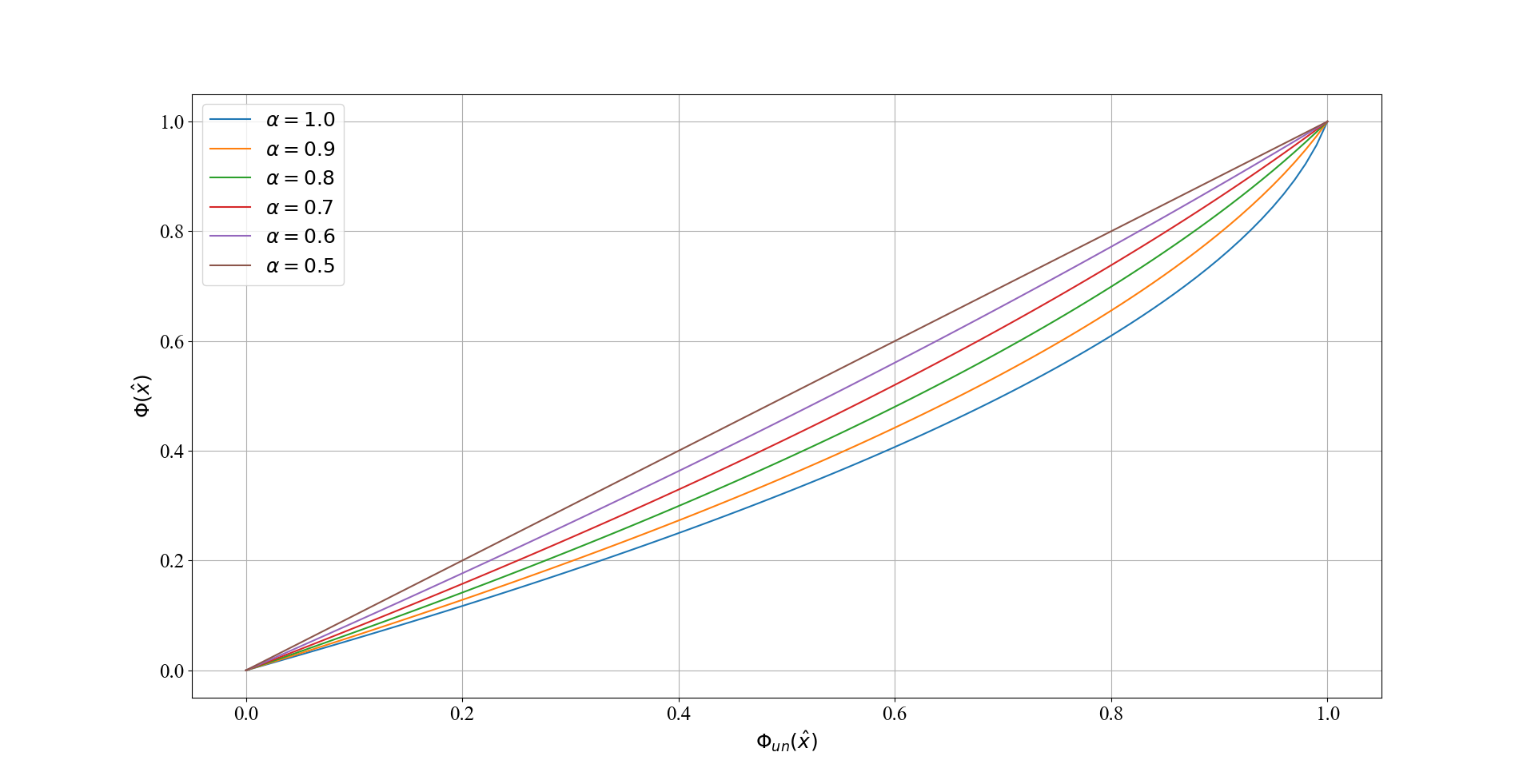}
	\caption{The transformation of two cumulative distribution functions $\Phi(\hat{x})$ and $\Phi_{\textsc{Un}}(\hat{x})$, where x-axis is the  C.D.F of unlabeled samples $\hat{\Phi}_{\textsc{Un}}$ , and y-axis is the anchor specific C.D.F $\Phi(\hat{x})$. From Eq~\eqref{eq:unlabel1}, it can be seen that when $\alpha=0.5$, which means the encoder makes random guesses, or when $\tau^+=0.5$, which means the prior class probabilities of positive and negative examples are equal, $\Phi(\hat{x})=\Phi_{\textsc{Un}}(\hat{x})$.}
	\label{fig:cdf_trans}
\end{figure*}

We applying total probability formula to relate marginal probabilities to conditional probabilities
\begin{eqnarray}
	\phi_{\textsc{Un}} &=& \tau^-\phi_{\textsc{Tn}} +\tau^+\phi_{\textsc{Fn}} \nonumber \\
	&=& 2\phi(\hat{x})[\alpha\tau^- + (1- \alpha)\tau^+  + (1-2\alpha)\Phi(\hat{x})(\tau^- -\tau^+)] \label{eq:unlabel1}
\end{eqnarray}
By integrating both sides of the Eq~\eqref{eq:unlabel1} we can establish a correlation between two distribution functions:
\begin{eqnarray}
	\Phi_{\textsc{Un}}(\hat{x})&=&\int_{-\infty}^{\hat x}  \phi_{\textsc{Un}}(t) dt \\
	&=& \int_{-\infty}^{\hat x} \tau^-\phi_{\textsc{Tn}}(t) +\tau^+\phi_{\textsc{Fn}}(t) dt \nonumber\\
	&=& \int_{-\infty}^{\hat x} 2\phi(t)[\alpha\tau^- + (1- \alpha)\tau^+  + (1-2\alpha)\Phi(t)(\tau^- -\tau^+)] dt\nonumber\\ 
	&=&  2[\alpha\tau^- + (1- \alpha)\tau^+]\int_{-\infty}^{\hat x} \phi(t) dt + (1-2\alpha)(\tau^- -\tau^+)\int_{-\infty}^{\hat x} 2 \phi(t) \Phi(t) dt \nonumber\\
	&=& 2[\alpha\tau^- + (1- \alpha)\tau^+] \Phi(\hat{x}) + (1-2\alpha)(\tau^- -\tau^+)\Phi^2(\hat{x})
\end{eqnarray}
Let 
\begin{eqnarray}
	a &=&  (1-2\alpha)(\tau^- -\tau^+)\nonumber \\
	b &=& 2[\alpha\tau^- + (1- \alpha)\tau^+] \nonumber
\end{eqnarray}
So we have $\Phi_{\textsc{Un}}(\hat{x})= a\Phi^2(\hat{x}) + b\Phi(\hat{x})$, and
\begin{eqnarray}
	\Phi(\hat{x}) = \frac{-b+\sqrt{b^2+4a\Phi_{\textsc{Un}}(\hat{x})}}{2a}
\end{eqnarray}
The other solution $\Phi(\hat{x}) = \frac{-b-\sqrt{b^2+4a\Phi_{\textsc{Un}}(\hat{x})}}{2a}$ is discarded because it is less than zero, which is not within the range of the cumulative distribution function. Fig~\ref{fig:cdf_trans} illustrates the transformation of two cumulative distribution functions $\Phi(\hat{x})$ and $\Phi_{\textsc{Un}}(\hat{x})$, where we fixed $\tau^+=0.1$. 
\subsection{Proof of Proposition~\ref{proposition:1}.}\label{Proof:ccd}
\begin{proposition}[Class Conditional Density]
	If $\phi(\hat x)$ is continuous density function that satisfy $\phi(\hat x) \geq 0 $ and $\int_{-\infty}^{+\infty}\phi(\hat x) d\hat x =1 $, then  $\phi_{\textsc{Tn}}(\hat{x})$ and $\phi_{\textsc{Fn}}(\hat{x})$ are probability density functions that satisfy $\phi_{\textsc{Tn}}(\hat{x})\geq 0$, $\phi_{\textsc{Fn}}(\hat{x})\geq 0$, and $\int_{-\infty}^{+\infty}\phi_{\textsc{Tn}}(\hat{x})d\hat x =1 $, $\int_{-\infty}^{+\infty}\phi_{\textsc{Fn}}(\hat{x})d\hat x =1 $.
	\begin{proof}
		Since $\phi(\hat x) \geq 0 $ and $\int_{-\infty}^{+\infty}\phi(\hat x) d\hat x =1 $, so
		\begin{eqnarray}
			\phi_{\textsc{Tn}}(\hat{x}) &=& \alpha \phi_{(1)}(\hat{x}) + (1-\alpha)\phi_{(2)}(\hat{x}) \nonumber \\ 
			&=& 2\alpha\phi(\hat{x}) [1-\Phi(\hat{x})]+ 2(1-\alpha)\phi(\hat{x}) \Phi(\hat{x}) \nonumber \\ 
			&\geq& 0,
		\end{eqnarray}
		where $\alpha \in [0.5,1]$ and $\Phi(\hat{x}) \in [0,1]$. 
		\begin{eqnarray}
			\int_{-\infty}^{+\infty}\phi_{\textsc{Tn}}(\hat{x})d\hat x &=& \int_{-\infty}^{+\infty}2\alpha\phi(\hat{x}) [1-\Phi(\hat{x})]+ 2(1-\alpha)\phi(\hat{x}) \Phi(\hat{x}) d\hat x \nonumber \\
			&=&2\alpha\int_{-\infty}^{+\infty}\phi(\hat{x}) [1-\Phi(\hat{x})] d\hat x+  2(1-\alpha)\int_{-\infty}^{+\infty} \phi(\hat{x}) \Phi(\hat{x}) d\hat x \nonumber \\
			&=&2\alpha\int_{-\infty}^{+\infty} [1-\Phi(\hat{x})] d\Phi(\hat{x}) +  2(1-\alpha)\int_{-\infty}^{+\infty}  \Phi(\hat{x}) d\Phi(\hat{x}) \nonumber \\
			&=& 2\alpha\int_{0}^{1} (1-\mu) d\mu +  2(1-\alpha)\int_{0}^{1}  \mu d\mu  \label{Eq:sub}\\
			&=& [\alpha(2\mu-\mu^2) +(1-\alpha)\mu^2] \big|_0^1 \nonumber \\
			&=&1,
		\end{eqnarray}
		where Eq.~\eqref{Eq:sub} is integration by substitution of $\Phi(\hat{x})$ and $\mu$. Likewise,	
		
		\begin{eqnarray}
			\phi_{\textsc{Fn}}(\hat{x}) &=& \alpha \phi_{(2)}(\hat{x}) + (1-\alpha)\phi_{(1)}(\hat{x}) \nonumber\\
			&=& 2\alpha \phi(\hat{x}) \Phi(\hat{x}) +2(1-\alpha)\phi(\hat{x}) [1-\Phi(\hat{x})] \nonumber \\ 
			&\geq& 0,
		\end{eqnarray}
		and 
		\begin{eqnarray}
			\int_{-\infty}^{+\infty}\phi_{\textsc{Fn}}(\hat{x})d\hat x &=& 2\alpha\int_{-\infty}^{+\infty}  \phi(\hat{x}) \Phi(\hat{x})d\hat x +2(1-\alpha)\int_{-\infty}^{+\infty}\phi(\hat{x}) [1-\Phi(\hat{x})] d\hat x \nonumber \\
			&=& 2\alpha\int_{-\infty}^{+\infty} \Phi(\hat{x})d\Phi(\hat{x}) +2(1-\alpha)\int_{-\infty}^{+\infty}[1-\Phi(\hat{x})] d\Phi(\hat{x}) \nonumber \\
			&=& 2\alpha\int_{0}^{1}  \mu d\mu +2(1-\alpha)\int_{0}^{1} (1-\mu) d\mu \nonumber \\
			&=& [\alpha \mu^2 + (1-\alpha) (2\mu - \mu^2) ]\big|_0^1 \nonumber \\
			&=&1.
		\end{eqnarray}
	\end{proof}
\end{proposition}

\begin{figure*}[h!]
	\centering
	\includegraphics[width=0.8\textwidth]{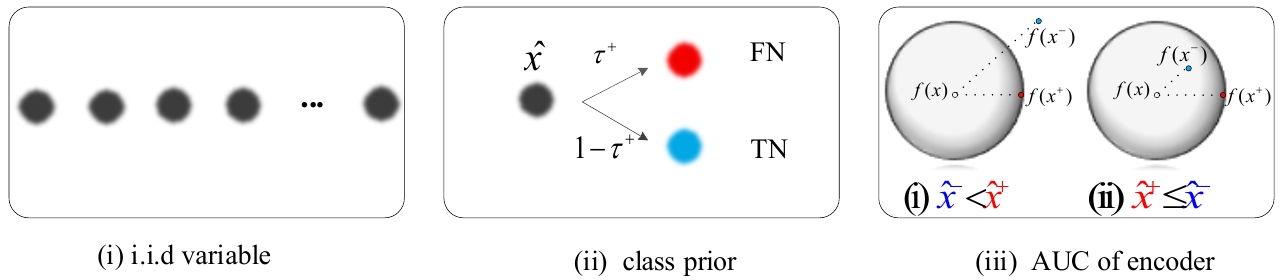}
	\caption{Three conditions.}
	\label{fig:condition}
\end{figure*}

The theoretical results of Lemma~\ref{lemma:1} to Lemma~\ref{lemma:3} are based on the following three lenient conditions as depicted in Fig~\ref{fig:condition}:
\begin{itemize}
 \item (i) Let the similarity score of unlabeled samples $\hat{x}$ be independently and identically distributed according to an unknown distribution $\phi$. 
 \item (ii) The class prior of positive samples $\hat{x}^+$ is $\tau^+$, and class prior of negative samples $\hat{x}^-$ is $\tau^- = 1- \tau^+$.
 \item (iii) $\mathbb{P}(\hat{x}^- < \hat{x}^+) = \alpha$, that is, the AUC of current encoder is $\alpha$. 
\end{itemize}
	
\subsection{Proof of Lemma~\ref{lemma:1}}\label{Proof:poster}	
\begin{lemma}[Posterior Probability Estimation]
	Under the aforementioned conditions, with $\beta = 0.5$, the  weight $\omega_i$ computed using Eq.~\eqref{eq:ome1} and the posterior probability of the sample being a true negative $\mathbb P(\textsc{Tn}|\hat{x}_i)$ satisfy
	\[\omega_i = \mathbb{P}(\textsc{Tn}|\hat{x}_i)/{\tau^-}\]
	
	\begin{proof}
		For independently and identically distributed random variable $\hat{x}$ with distribution $\phi$, the distributions of the order statistics~\cite{David:2004} $X_{(1)}$ and $X_{(2)}$ are given by:
		\begin{eqnarray}
			\phi_{(1)}(\hat{x}) &=& 2\phi(\hat{x}) [1-\Phi(\hat{x})] \nonumber \\
			\phi_{(2)}(\hat{x}) &=& 2\phi(\hat{x}) \Phi(\hat{x}), \nonumber
		\end{eqnarray}
		where $\Phi(\hat{x}) = \int_{-\infty}^{\hat{x}} \phi(t) dt$ is the cumulative distribution function corresponding to the probability density function $\phi(\hat{x})$.
		
		Since $\mathbb{P}(\hat{x}^- < \hat{x}^+) = \alpha$, so 
		\[\mathbb{P}(\hat{x}^- \geq \hat{x}^+) = 1- \alpha.\]
		
		Then, the probability density of the negative class $\hat{x}^-$ in $\hat{x}$ is a mixture of the probability densities of the order statistics $X_{(1)}$ and $X_{(2)}$ with coefficient $\alpha$:
		\begin{eqnarray}
			\phi_{\textsc{Tn}}(\hat{x}) = \alpha \phi_{(1)}(\hat{x}) + (1-\alpha)\phi_{(2)}(\hat{x}), \nonumber
		\end{eqnarray}
		where $\phi_{\textsc{Tn}}(\hat{x})$ represents the class-conditional probability density of the negative class. Similarly, the class-conditional probability density of the positive class $\hat{x}^+$ is given by:
		\begin{eqnarray}
			\phi_{\textsc{Fn}}(\hat{x}) = \alpha \phi_{(2)}(\hat{x}) + (1-\alpha)\phi_{(1)}(\hat{x}),\nonumber
		\end{eqnarray}
		Then, the probability density of the observed unlabeled sample is:
		\begin{eqnarray}
			\phi_{\textsc{Un}}(\hat{x}) &=& \tau^-\phi_{\textsc{Tn}}(\hat{x}) +\tau^+\phi_{\textsc{Fn}}(\hat{x}) \nonumber
		\end{eqnarray}
		where $\tau^- = 1 - \tau^+$ represents the proportion of the negative class. The above conclusions have been stated in the method section and will not be further elaborated. Then, the $\omega_i(\hat{x}_i;\alpha,\beta=0.5)$ in Equation \eqref{eq:ome} can be computed as:
		\begin{eqnarray}
			\phi_\textsc{Thn}(\hat{x}; \alpha, \beta=0.5)  &=&\frac{ (1-0.5)\alpha  \phi_{(1)}(\hat{x}) +0.5 (1-\alpha)  \phi_{(2)}(\hat{x}) }{(1-0.5)\alpha + 0.5 (1-\alpha)} \label{eq:phi_htnbeta}\nonumber\\
			&=& \alpha  \phi_{(1)}(\hat{x})+(1-\alpha)  \phi_{(2)}(\hat{x})\nonumber\\
			&=& \phi_\textsc{Tn}(\hat{x}). \label{eq:phi_htnbeta5}
		\end{eqnarray}
		Inserting Eq~\eqref{eq:phi_htnbeta5} back to Eq~\eqref{eq:ome}, we obtain
		\begin{eqnarray}
			\omega(\hat x_i) 
			&=& \frac{\phi_\textsc{Tn}(\hat{x}_i)}{\phi_{\textsc{Un}}(\hat{x}_i)} \nonumber \\
			&=& \frac{\phi_\textsc{Tn}(\hat{x}_i) \tau^-}{\phi_{\textsc{Tn}}(\hat{x}_i)\tau^-+\phi_{\textsc{Fn}}(\hat{x}_i)\tau^+} \frac{1}{\tau^-} \nonumber \\
			&=& p(\textsc{Tn}|\hat{x}_i) \frac{1}{\tau^-},
		\end{eqnarray}
		which completes the proof. More specifically, the posterior probability estimation can be expressed as a function of the cumulative distribution function (C.D.F) $\Phi(\hat{x}_i)$ and the prior probability $\tau$ as follows:
		\begin{eqnarray}
			p(\textsc{Tn}|\hat{x}_i) 
			&=& \frac{\phi_{\textsc{Tn}}(\hat{x}_i)\tau^-}{\phi_{\textsc{Tn}}(\hat{x}_i)\tau^- + \phi_{\textsc{Fn}}(\hat{x}_i)\tau^+} \nonumber\\
			&=& \frac{\alpha\tau^- + (1-2\alpha) \Phi(\hat{x}_i)\tau^-}
			{\alpha\tau^- + (1- \alpha)\tau^+  + (1-2\alpha)\Phi(\hat{x}_i)(\tau^- -\tau^+)} \label{Eq:poster}
		\end{eqnarray}
		It is worth noting that the unknown distribution of unlabeled samples $\phi(\hat{x}_i)$ has been eliminated. Although the explicit expression of $\Phi(\hat{x}_i)=\int_{-\infty}^{\hat{x}_i} \phi(t)dt$ is unknown, we can calculate the empirical C.D.F instead. Therefore, the posterior probability $p(\textsc{Tn}|\hat{x}_i)$ can be analytically calculated without making any parametric assumptions about $\phi(\hat{x}_i)$. 
	\end{proof}
\end{lemma}

\subsection{Proof of Lemma~\ref{lemma:2}}\label{Proof:unbias}
\begin{lemma}[Asymptotic Unbiased Estimation]
	Under the aforementioned conditions, with $\beta = 0.5$, for any encoder $f$, as $N\rightarrow \infty$, we have
	\[\mathcal{L}_\textsc{~Bcl}\rightarrow\mathcal{L}_\textsc{~Sup}\]
	\begin{proof}
		We draw the conclusion by the Lebesgue Dominant Convergence Theorem and the properties of important sampling:
		\begin{eqnarray}
			\lim_{N\rightarrow \infty} \mathcal{L}_\textsc{~Bcl} &=& \lim_{N\rightarrow \infty}   \mathbb E_{\substack{x \sim p_d\\~x^+ \sim p^+\\~x^- \sim p_d}} -\log[ \frac{e^{f(x)^Tf(x^+)}}{e^{f(x)^Tf(x^+)} +  \sum_{i=1}^N  \omega_i\cdot e^{f(x)^Tf(x^-_i)} }] \nonumber\\ 
			&=&   \mathbb E_{\substack{x \sim p_d\\~x^+ \sim p^+\\~x^- \sim p_d}} \lim_{N\rightarrow \infty} -\log[ \frac{e^{f(x)^Tf(x^+)}}{e^{f(x)^Tf(x^+)} +  \sum_{i=1}^N  \omega_i\cdot e^{f(x)^Tf(x^-_i)} }] \nonumber\\ 
			&=& \mathbb E_{\substack{x \sim p_d\\~x^+ \sim p^+} }	\lim_{N\rightarrow \infty}[-\log \frac{e^{f(x)^Tf(x^+)}}{e^{f(x)^Tf(x^+)} + N\cdot \mathbb{E}_{\hat{x} \sim \phi_{\textsc{Un}} } \omega\hat x  }] \label{eq:bcldenomitor}
		\end{eqnarray}
		where \[\omega(\hat x;\alpha, \beta) = \frac{\phi_\textsc{Thn}(\hat{x};\alpha,\beta)}{\phi_{\textsc{Un}}(\hat{x})}
		\]
		when taking $\beta=0.5$,
		\begin{eqnarray}
			\phi_\textsc{Thn}(\hat{x}; \alpha, \beta=0.5)  &=&\frac{ (1-0.5)\alpha  \phi_{(1)}(\hat{x}) +0.5 (1-\alpha)  \phi_{(2)}(\hat{x}) }{(1-0.5)\alpha + 0.5 (1-\alpha)} \label{eq:phi_htnbeta}\nonumber\\
			&=& \alpha  \phi_{(1)}(\hat{x})+(1-\alpha)  \phi_{(2)}(\hat{x})\nonumber\\
			&=& \phi_\textsc{Tn}(\hat{x}).
		\end{eqnarray}
		Therefor, $\omega=\frac{\phi_\textsc{Tn}(\hat{x};\alpha,\beta)}{\phi_{\textsc{Un}}(\hat{x})}$, so the second term in the denominator of Eq.~\eqref{eq:bcldenomitor}
		\begin{eqnarray}
			\mathbb{E}_ {\hat{x} \sim \phi_{\textsc{Un}}} \omega \hat{x} &=& \int \omega \hat{x}\phi_{\textsc{Un}}(\hat{x}) d\hat{x} \nonumber\\
			&=& \int \hat{x} \frac{\phi_\textsc{Tn}(\hat{x};\alpha,\beta)}{\phi_{\textsc{Un}}(\hat{x})}  \phi_{\textsc{Un}}(\hat{x}) d\hat{x} \nonumber \\
			&=& \int  \hat{x}\phi_\textsc{Tn}(\hat{x})d\hat{x} \nonumber\\
			&=& \mathbb{E}_ {x^-\sim p^-} e^{f(x)^Tf(x_j^-)}\label{eq:sumtn}
		\end{eqnarray}
		Inserting Eq.~\eqref{eq:sumtn} back to Eq.~\eqref{eq:bcldenomitor}, we obtain
		\begin{eqnarray}
			\lim_{N\rightarrow \infty} \mathcal{L}_\textsc{~Bcl} &=& \mathbb E_{\substack{x \sim p_d\\~x^+ \sim p^+} }	\lim_{N\rightarrow \infty}[-\log \frac{e^{f(x)^Tf(x^+)}}{e^{f(x)^Tf(x^+)} +  N\mathbb{E}_ {x^-\sim p^-} e^{f(x)^Tf(x_j^-)}}] \nonumber\\
			&=& \mathbb E_{\substack{x \sim p_d\\~x^+ \sim p^+\\~x^- \sim p^-} }	\lim_{N\rightarrow \infty}[-\log \frac{e^{f(x)^Tf(x^+)}}{e^{f(x)^Tf(x^+)} + \sum_{j=1}^Ne^{f(x)^Tf(x_j^-)}}] \nonumber\\
			&=& \lim_{N\rightarrow \infty} \mathbb E_{\substack{x \sim p_d\\~x^+ \sim p^+\\~x^- \sim p^-} }	[-\log \frac{e^{f(x)^Tf(x^+)}}{e^{f(x)^Tf(x^+)} +  \sum_{j=1}^Ne^{f(x)^Tf(x_j^-)} }] \label{eq:dominate}\\
			&=& \lim_{N\rightarrow \infty} \mathcal{L}_\textsc{~Sup}
		\end{eqnarray}
		where Eq.~\eqref{eq:dominate} is obtained by Lebesgue Dominant Convergence Theorem.
	\end{proof}
\end{lemma}

\subsection{Proof of Lemma~\ref{lemma:3}}\label{Proof:bound}
\begin{lemma}[Estimation Error Bound]
	Under the aforementioned conditions, with $\beta = 0.5$, for any encoder $f$ and $N> 0$, 
	\[|\hat{\mathcal{L}}_\textsc{~Bcl}-\mathcal{L}^{~\text{N}}_\textsc{~Bcl}|\leq \tau^- \sqrt{\frac{2\pi}{N}}\]
	\begin{proof}
		The similarity score for the true negative samples is denoted as $\theta =\mathbb{E}_{x^-\sim p^-} \hat{x}^-$, and the BCL estimator using unlabeled samples' similarity scores is denoted as $\hat{\theta}_\textsc{Bcl} = \frac{1}{N} \sum_{i=1}^{N} \omega_i \hat{x}_i^\prime$.
		
		Let anchor $x$ and positive sample $x^+$ be fixed, we denote the following quantity 
		\begin{eqnarray}
			\triangle = |-\log \frac{\hat{x}^+}{\hat{x}^+ + N \hat\theta} +\log \frac{\hat{x}^+}{\hat{x}^+ + N \theta} |
		\end{eqnarray} 
		
		Applying the intermediate results of Lemma A.2. of the DCL\cite{Chuang:2020:NIPS} paper, we have
		\begin{eqnarray}
			\mathbb{P}(\triangle \geq \epsilon) &\leq& \mathbb{P}(|\hat \theta_\textsc{Bcl} -\theta|\geq \epsilon e^{-1})  \label{eq:lemma43} 
		\end{eqnarray}
		
		Then we seek to bound the differences of $\hat{\theta}_\textsc{BCL}$  by substituting the value of the $i$-th random variable $\hat{x}_i^\prime$ with 
		$\hat{x}_i^{\prime\prime}$. Let 
		\[h: h(\hat{x}^\prime_1,\hat{x}^\prime_2,\cdots, \hat{x}^\prime_N) \rightarrow \frac{1}{N} \sum_{i=1}^{N} \omega_i \hat{x}_i^\prime\]
		Since the embedding $f(x^\prime)$ lies on the surface of the unit sphere, we have $f(x)^Tf(x_i^\prime) \in [-1,1]$, thus $\hat{x}_i^\prime = \exp(f(x)^Tf(x_i^\prime)) \in [-1/e,1/e]$. So 
		\begin{eqnarray}
			\sup |h(\hat{x}_1\prime, \cdots, \hat{x}_i\prime,\cdots \hat{x}_N\prime)-h(\hat{x}_1\prime, \cdots, \hat{x}_i{\prime\prime},\cdots \hat{x}_N\prime)| 
			&\leq& \frac{2}{Ne}\omega_i \label{eq:lemma41} \\
			&\leq& \frac{2}{Ne\tau^-}\label{eq:lemma42}
		\end{eqnarray}
		where Eq.\eqref{eq:lemma41} is obtained since $|\hat{x}^\prime_i-\hat{x}_i^{\prime\prime}| \leq \frac{2}{e}$, and Eq.\eqref{eq:lemma42} is obtained since $\omega_i = p(\textsc{Tn}|\hat{x}_i) \frac{1}{\tau^-} \leq \frac{1}{\tau^-}$.
		
		McDiarmid's Inequality states that
		if ${\displaystyle h:{\mathcal {X}}_{1}\times {\mathcal {X}}_{2}\times \cdots \times {\mathcal {X}}_{n}\rightarrow \mathbb {R} } $ satisfy the bounded differences property with bounds $
		{\displaystyle c_{1},c_{2},\dots ,c_{n}}$, then for independent random variables $X_{1},X_{2},\dots ,X_{n}$, where 
		${\displaystyle X_{i}\in {\mathcal {X}}_{i}}$ for all $i$. Then, for any $\epsilon \geq 0$,  
		\[{\displaystyle {\text{P}}(|h(X_{1},X_{2},\ldots ,X_{n})-\mathbb {E} [h(X_{1},X_{2},\ldots ,X_{n})]|\geq \epsilon )\leq 2\exp \left(-{\frac {2\epsilon ^{2}}{\sum _{i=1}^{n}c_{i}^{2}}}\right).}\]
		
		Taking $c_1=c_2=\cdots =c_n = \frac{2}{Ne\tau^-}$ and applying the McDiarmid's Inequality, we obtain
		\begin{eqnarray}
			\mathbb{P}(|\hat \theta_{\textsc{Bcl}} -\theta|\geq \epsilon e^{-1}) 
			&=& \mathbb{P}(|h(\hat{x}_1\prime, \cdots, \hat{x}_i\prime,\cdots \hat{x}_N\prime) - \mathbb{E} ~h(\hat{x}_1\prime, \cdots, \hat{x}_i\prime,\cdots \hat{x}_N\prime)|\geq \epsilon e^{-1} ) \\
			&\leq& 2 \exp(-\frac{N\epsilon^2}{2(\tau^-)^2}) \label{eq:lemma44}
		\end{eqnarray}
		Inserting Eq.~\eqref{eq:lemma44} back to Eq.~\eqref{eq:lemma43}, we obtain
		\begin{eqnarray}
			\mathbb{P}(\triangle \geq \epsilon) \leq  2 \exp(-\frac{N\epsilon^2}{2(\tau^-)^2}). \label{eq:lemma45}
		\end{eqnarray}

		So
		\begin{eqnarray}
			|\hat{\mathcal{L}}_\textsc{~Bcl}-\mathcal{L}^{~N}_\textsc{~Bcl}| &=&  |-\mathbb{E}_{x,x^+} \log \frac{\hat{x}^+}{\hat{x}^+ + N \hat\theta} + \mathbb{E}_{x,x^+}  \log \frac{\hat{x}^+}{\hat{x}^+ + N \theta} | \nonumber \\
			&\leq&\mathbb{E}_{x,x^+}  |-\log \frac{\hat{x}^+}{\hat{x}^+ + N \hat\theta} +\log \frac{\hat{x}^+}{\hat{x}^+ + N \theta} | \label{eq:lemma46}\\
			&=& \mathbb{E}_{x,x^+}  \triangle \nonumber\\
			&=& \mathbb{E}_{x,x^+}  \left[\int_{0}^{\infty} \mathbb{P}(\triangle \geq \epsilon|x,x^+)d \epsilon \right] \nonumber \\
			&\leq& \int_{0}^{\infty}2 \exp(-\frac{N\epsilon^2}{2(\tau^-)^2}) d\epsilon \nonumber\\
			&=&\tau^- \sqrt{\frac{2\pi}{N}} \label{eq:lemma47},
		\end{eqnarray}
where Eq.~\eqref{eq:lemma46} is obtained by applying Jensen's inequality to push the absolute value inside the expectation, and Eq.~\eqref{eq:lemma47} is obtained by using the Gaussian integral.
	\end{proof}
\end{lemma}
\section{Numerical Experiments}\label{Sec:numer}

\begin{algorithm}[!]
	\caption{Numerical experiments.}\label{Alg:numer}
	\KwIn{location parameter $\alpha$(mixtrue coffecient), temperature scaling $t$ }
	\KwOut{Observations and labels.}
	\BlankLine
	\For{\textsf{anchor} $m=1, 2, ..., M$}{
		~~$\backslash$$\backslash$ \textit{choose an anchor-specific distribution $\phi$. } \\
		Random select $[a,b] \subseteq [-1/t^2, 1/t^2]$\\
		Set $\phi\sim U(a,b) $ \\
		\For{\textsf{negative} $ j=1, 2, ..., N$}{
			~~$\backslash$$\backslash$ \textit{step1:selecting one population} \\
			~~$p$ = random.uniform(0,1)\\
			\If{$p \leq \tau^+$}{~~$\backslash$$\backslash$ \textit{step2:generating observation from $\phi_{\textsc{Fn}}$} \\
				~~$\hat{x}_j$ = AccRejetSamplingFN($\phi_{\textsc{Fn}}$)\\
				label = False
			}
			\Else{~~$\backslash$$\backslash$ \textit{step2:generating observation from $\phi_{\textsc{Tn}}$}\\
				~~$\hat{x}_j$ = AccRejetSamplingTN($\phi_{\textsc{Tn}}$) \\
				label = True
			}
			~~$\backslash$$\backslash$ \textit{step3:mapping} \\
			$\hat{x}_j = \exp(\hat{x}_j/t)$ \\
			collecting observation $\hat{x}$ and label
		}
	}
	
	\KwResult{Observations and labels.}
\end{algorithm}

\subsection{Implementation Details of Accept-Reject Sampling} \label{appb:accrej}
The objective is to generate sample $\hat{x}$ from class conditional density $\phi_{\textsc{Tn}}(\hat{x})$, i.e., $\hat{x} \sim \phi_{\textsc{Tn}}(\hat{x})$. The basic idea is to generate a sample $\hat{x}$ from proposal distribution $\phi$, and accept it with acceptance probability $p_{\textsc{Tn}}$. To calculate the acceptance probability $p_{\textsc{Tn}}$, we first write $\phi_{\textsc{Tn}}(\hat{x})$ as a function of proposal distribution $\phi(\hat{x})$
\begin{eqnarray}
	\phi_{\textsc{Tn}}(\hat{x}) &=& \alpha \phi_{(1)}(\hat{x}) + (1-\alpha)\phi_{(2)}(\hat{x}) \nonumber\\
	&=& [2\alpha+(2-4\alpha)\Phi(\hat{x})]\phi(\hat x), \nonumber
\end{eqnarray}
where $\Phi(\hat{x})$ is the c.d.f. of proposal distribution $\phi(\hat{x})$. Next we find a minimum $c$ that satisfies the following inequality:
\[c \cdot \phi(\hat{x}) \geq \phi_{\textsc{Tn}}(\hat{x}),\]
that is
\[c \cdot \phi(\hat{x}) \geq  [2\alpha+(2-4\alpha)\Phi(\hat{x})]\phi(\hat x).\]
So the minimal of $c$ is attained
\begin{eqnarray}
	c &=& min [2\alpha+(2-4\alpha)\Phi(\hat{x})], \Phi(\hat{x}) \in [0,1]\nonumber\\
	&=&2\alpha, \nonumber
\end{eqnarray}
when $\Phi(\hat{x})=0$ since $2-4\alpha \leq 0$. So the acceptance probability is
\begin{eqnarray}
	p_{\textsc{Tn}} &=& \frac{\phi_{\textsc{Tn}}(\hat{x})}{c \cdot \phi(\hat{x}) } \nonumber \\
	&=& [\alpha + (1-2 \alpha)\cdot \Phi(\hat{x})] /\alpha \nonumber
\end{eqnarray}
An observation $\hat{x}$ generated from proposal distribution  $\phi(\hat{x})$ is accepted with probability $p_{\textsc{Tn}} $, formulates the empirical observations $\hat{x} \sim \phi_{\textsc{Tn}}(\hat{x})$ as described in Algorithm~\ref{Alg:AccRetTN}.
\begin{algorithm}[!]
	\caption{~AccRejetSamplingTN($\phi_{\textsc{Tn}}$).}\label{Alg:AccRetTN}
	\KwIn{location parameter $\alpha$, proposal distrition $\phi$ }
	\KwOut{samples $\hat{x} \sim \phi_{\textsc{Tn}}$.}
	\BlankLine
	~~$\hat{x}_j$ = generate observation $\hat{x}_j$ from $\phi$ \\
	cdf = $\int_{-\infty}^{\hat{x}_j} \phi(t)dt$ \\
	u = random.uniform$(0,1)$ \\
	\While{u $ > [\alpha + (1-2 \alpha)\cdot cdf] /\alpha ~$~ }{
		~~$\hat{x}_j$ = generate observation $\hat{x}_j$ from $\phi$ \\
		cdf = $\int_{-\infty}^{\hat{x}_j} \phi(t)dt$ \\
		u = random.uniform$(0,1)$
	}
	
	\KwResult{$\hat{x}_j$}
\end{algorithm}

Likewise, the acceptance probability for $\phi_{\textsc{Fn}}$
\[p_{\textsc{Fn}} = [1 - \alpha + (2 \alpha - 1)\cdot \Phi(\hat{x})] / \alpha \]
can be calculated in the similar way. An observation $\hat{x}$ from proposal distribution  $\phi(\hat{x})$ is accepted with probability $p_{\textsc{Fn}} $, formulates the empirical observations $\hat{x} \sim \phi_{\textsc{Fn}}(\hat{x})$ as described in Algorithm~\ref{Alg:AccRetFN}.
\begin{algorithm}[!]
	\caption{~AccRejetSamplingFN($\phi_{\textsc{Fn}}$).}\label{Alg:AccRetFN}
	\KwIn{location parameter $\alpha$, proposal distrition $\phi$ }
	\KwOut{samples $\hat{x} \sim \phi_{\textsc{Fn}}$.}
	\BlankLine
	~~$\hat{x}_j$ = generate observation $\hat{x}_j$ from $\phi$ \\
	cdf = $\int_{-\infty}^{\hat{x}_j} \phi(t)dt$ \\
	u = random.uniform$(0,1)$ \\
	\While{u $ > [1 - \alpha + (2 \alpha - 1)\cdot cdf ] /\alpha ~$~ }{
		~~$\hat{x}_j$ = generate observation $\hat{x}_j$ from $\phi$ \\
		cdf = $\int_{-\infty}^{\hat{x}_j} \phi(t)dt$ \\
		u = random.uniform$(0,1)$
	}
	
	\KwResult{$\hat{x}_j$}
\end{algorithm}

\subsection{Parameter Settings}\label{appb:para}
For baseline estimator $\hat{\theta}_\textsc{Dcl}$ in Eq~\eqref{Eq:DCLEstimator}, the number of positive samples $K$ is fixed as 10. We start from fixing the parameters as: $\alpha=0.9$, $\beta=0.5$, $\gamma=0.1$ $\tau^+=0.1$, temperature scaling $t=0.5$, number of anchors $M=1e3$, and number of negative samples $N=64$ to investigate the performance of different estimators under different parameters settings. 

\begin{figure*}[t]
	\centering
	\subfigure[Performance of encoder $f$.]
	{\includegraphics[width=0.32\textwidth, height=0.2\hsize]{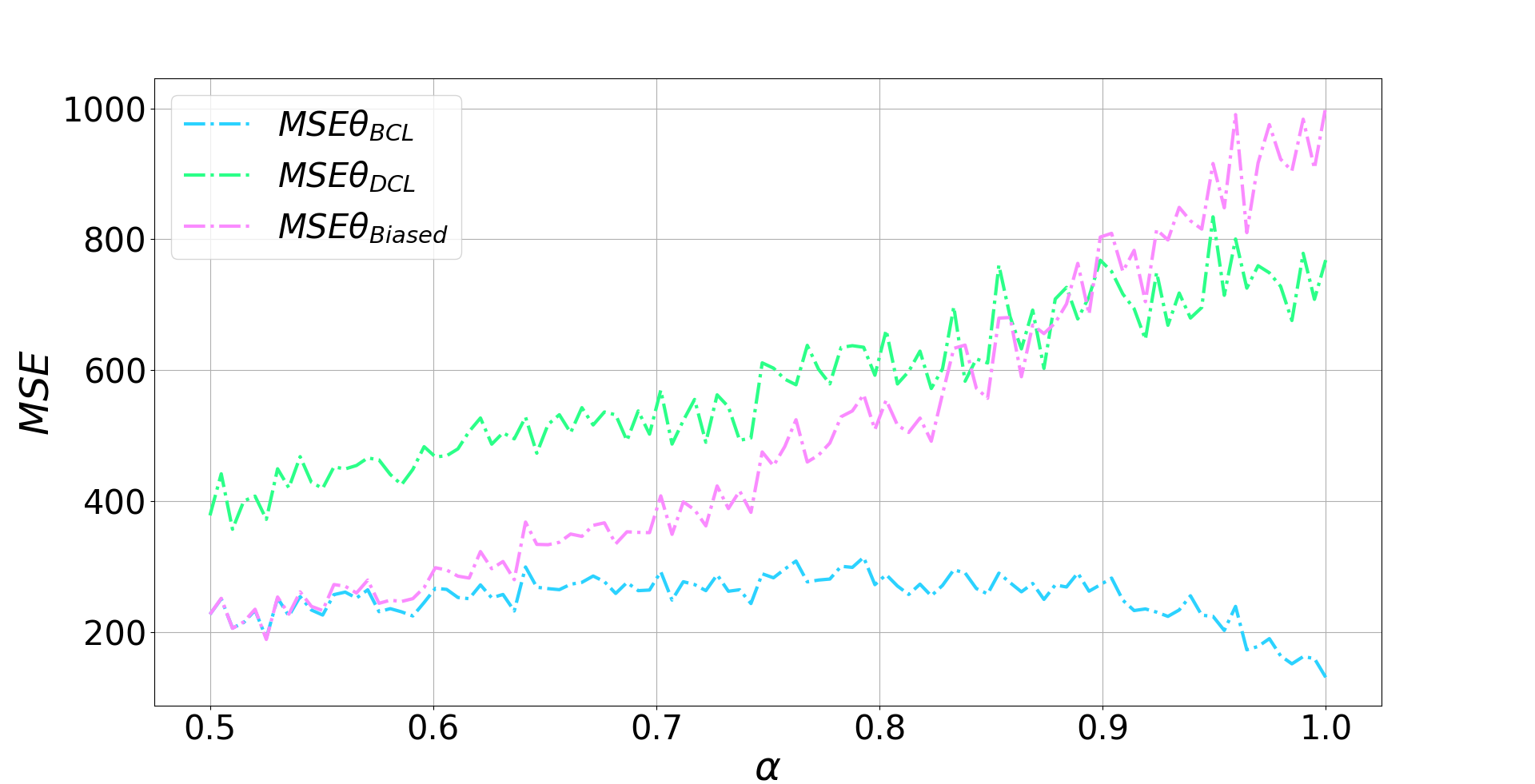}\label{fig:msealpha}}
	\subfigure[Number of negative samples]
	{\includegraphics[width=0.32\textwidth, height=0.2\hsize]{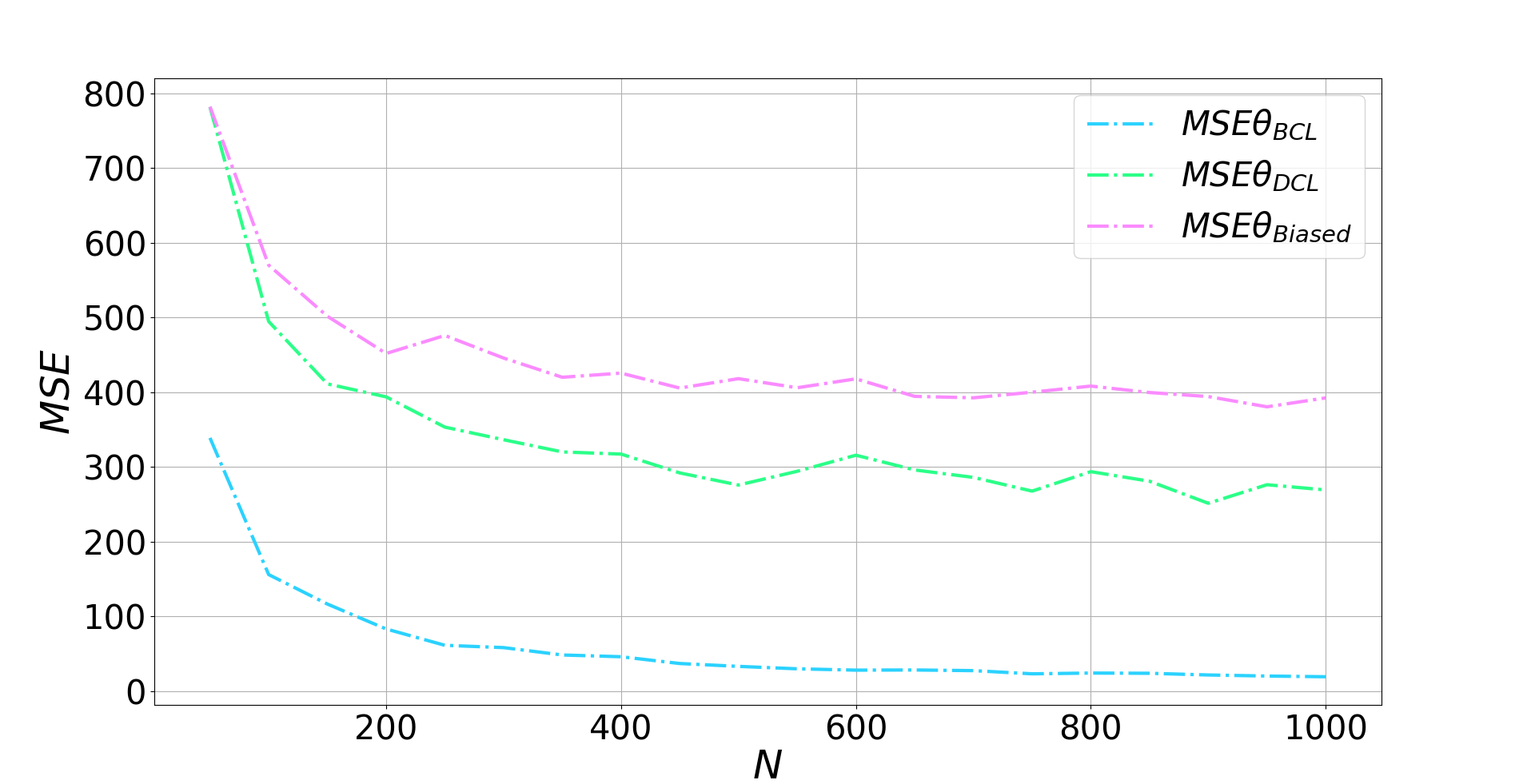}\label{fig:mseN}}
	\subfigure[Class probability.]
	{\includegraphics[width=0.32\textwidth, height=0.2\hsize]{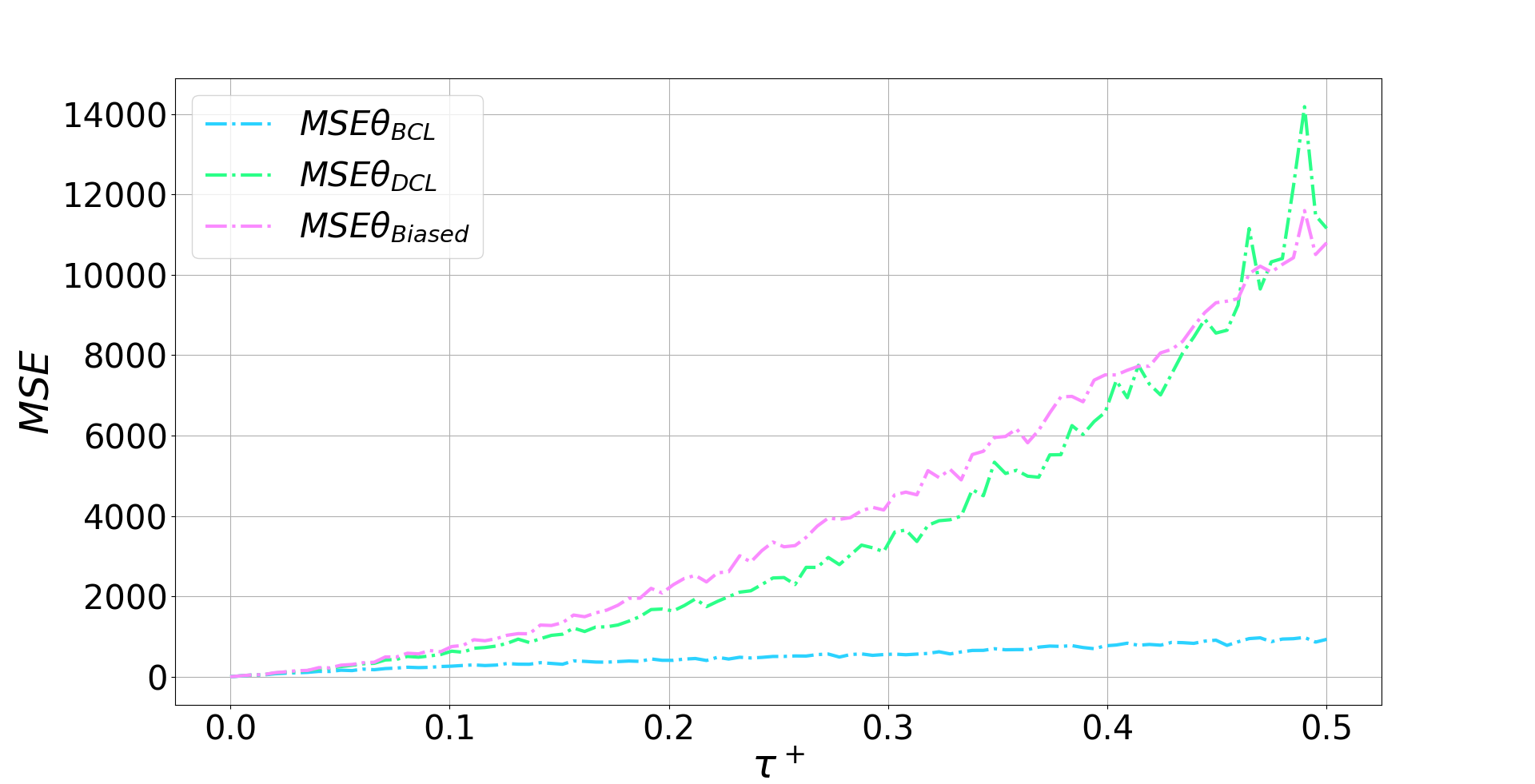}\label{fig:msetau}}
	\subfigure[Variation of proposal distribution $\phi$.]
	{\includegraphics[width=0.32\textwidth, height=0.2\hsize]{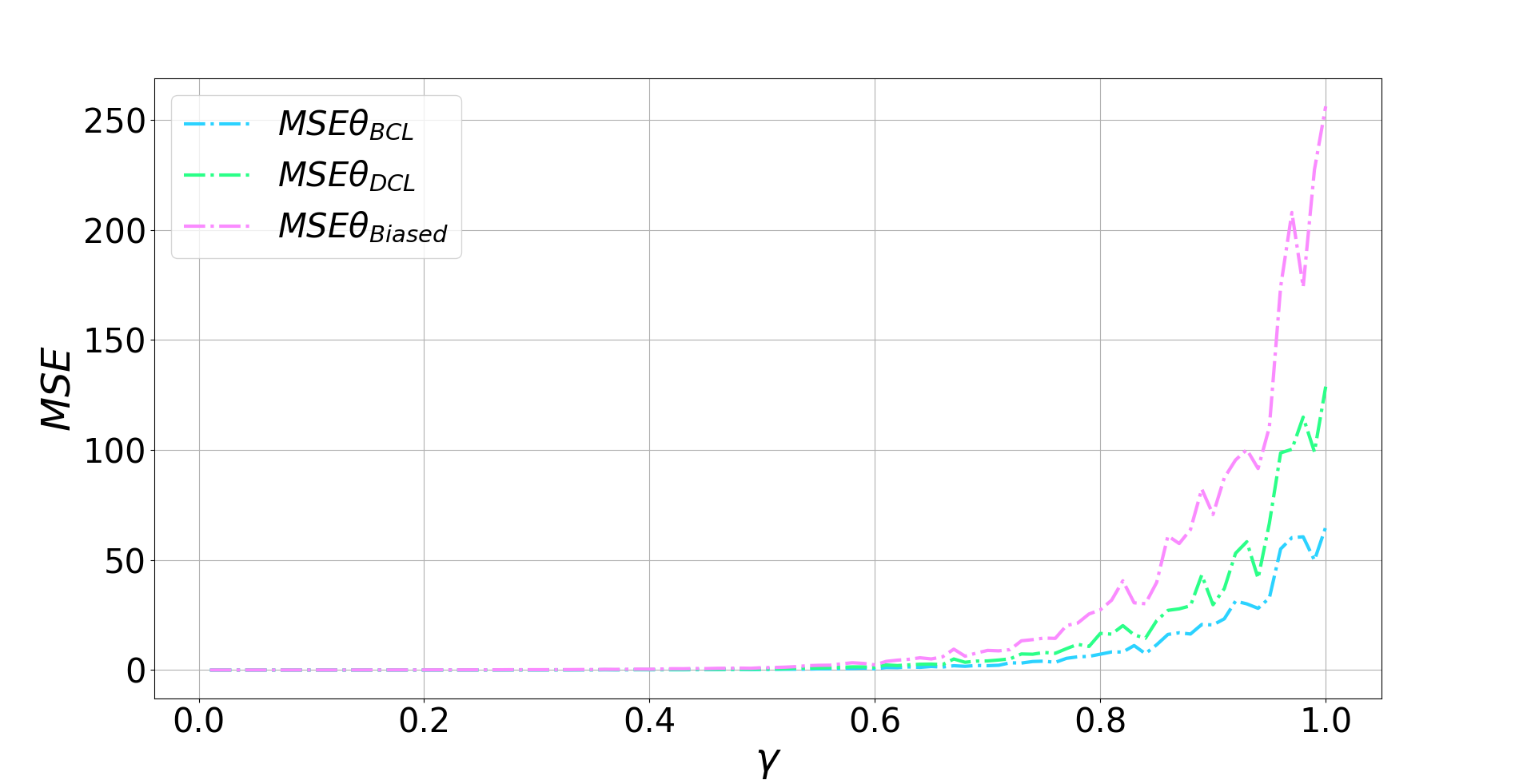}\label{fig:msegamma}}
	\subfigure[Temperature scaling.]
	{\includegraphics[width=0.32\textwidth, height=0.2\hsize]{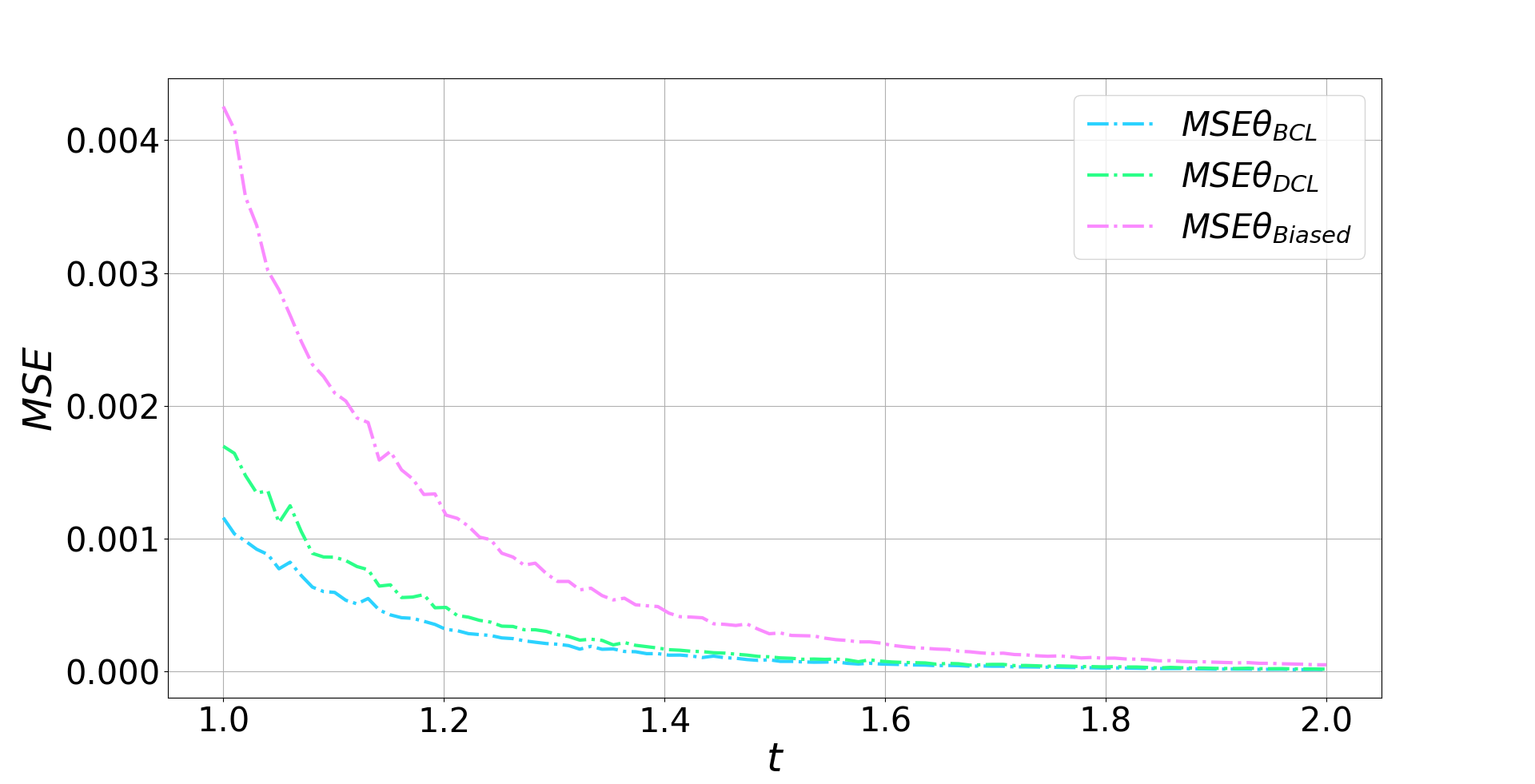}\label{fig:mset}}
	\subfigure[Number of anchors.]
	{\includegraphics[width=0.32\textwidth, height=0.2\hsize]{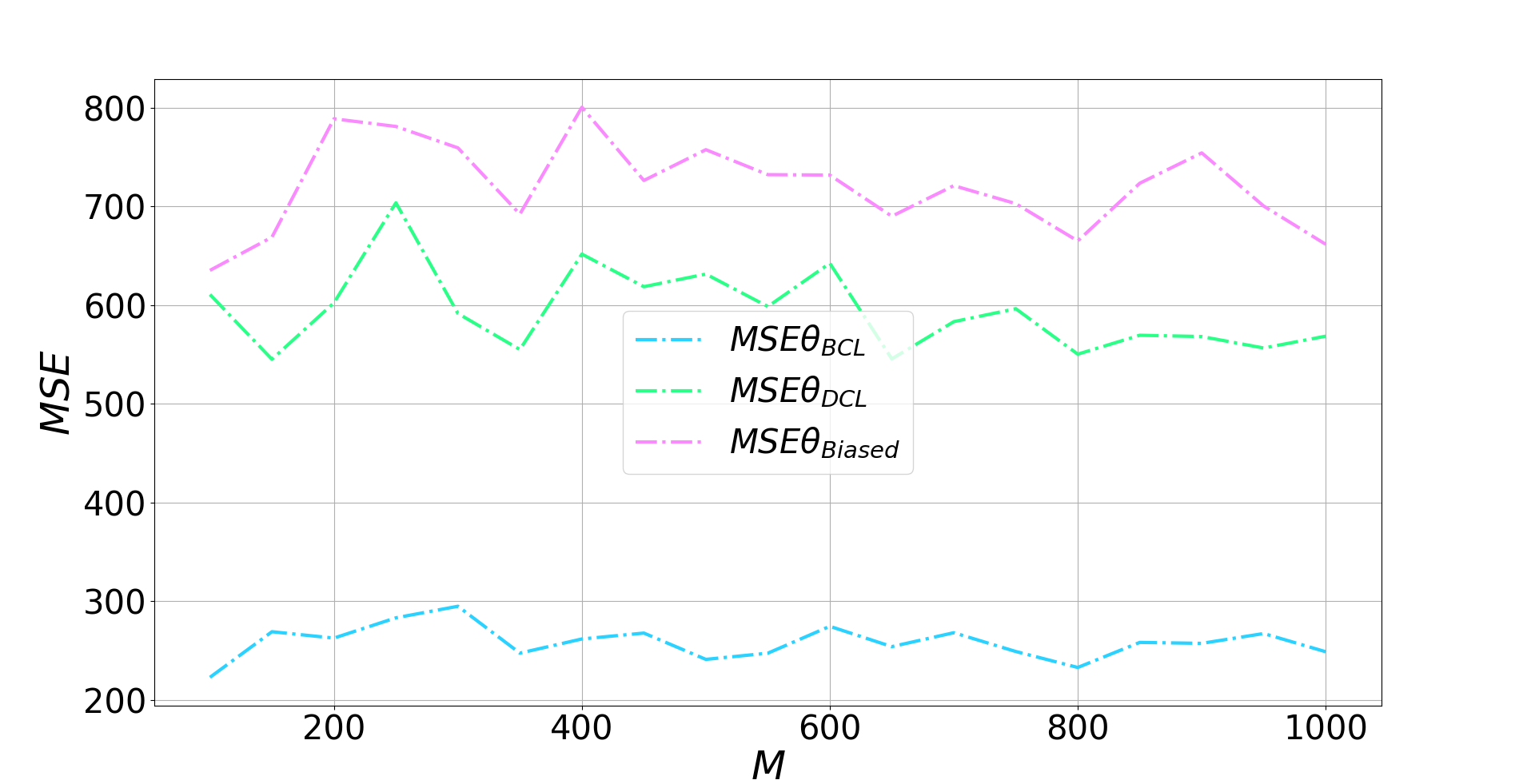}\label{fig:mseM}}
	\caption{$\textsc{Mse}$ of different estimators with various parameter settings.}
	\label{Fig:MSE}
\end{figure*}

A particular notice here is that we set $\beta=0.5$ in the $\omega$ computation, as hard negative mining parameter $\beta$ is designed in consideration of the requirements of downstream task that pushing $\textsc{Tn}$ samples further apart, other than statistical quality of $\hat{\theta}_{\textsc{Bcl}}$. Additionally, an important comparative method, HCL\cite{Robinson:2021:ICLR}, was not included in the numerical experiments. This is because HCL is primarily designed for the task of hard negative mining rather than considering the consistency of the estimator. Due to the up-weighting of difficult negative samples, HCL exhibits exponential growth of mean squared error (MSE). The results are presented in Fig~\ref{Fig:MSE}.

\begin{figure}[!]
	\centering
	\subfigure[]
	{\includegraphics[width=0.49\textwidth, height=0.3\hsize]{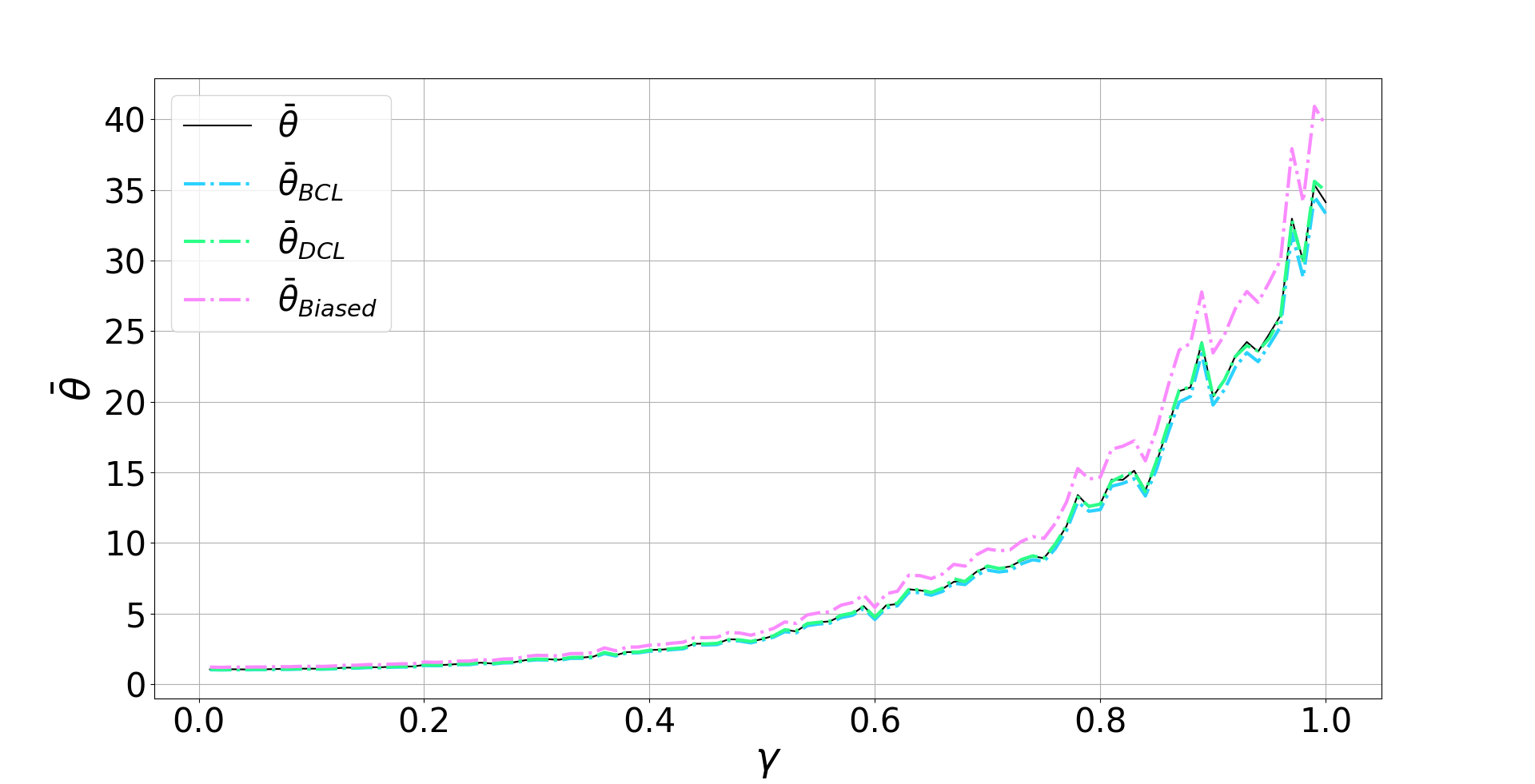}\label{fig: numerbar_gamma}}\hspace{0.1cm}
	\subfigure[]
	{\includegraphics[width=0.49\textwidth, height=0.3\hsize]{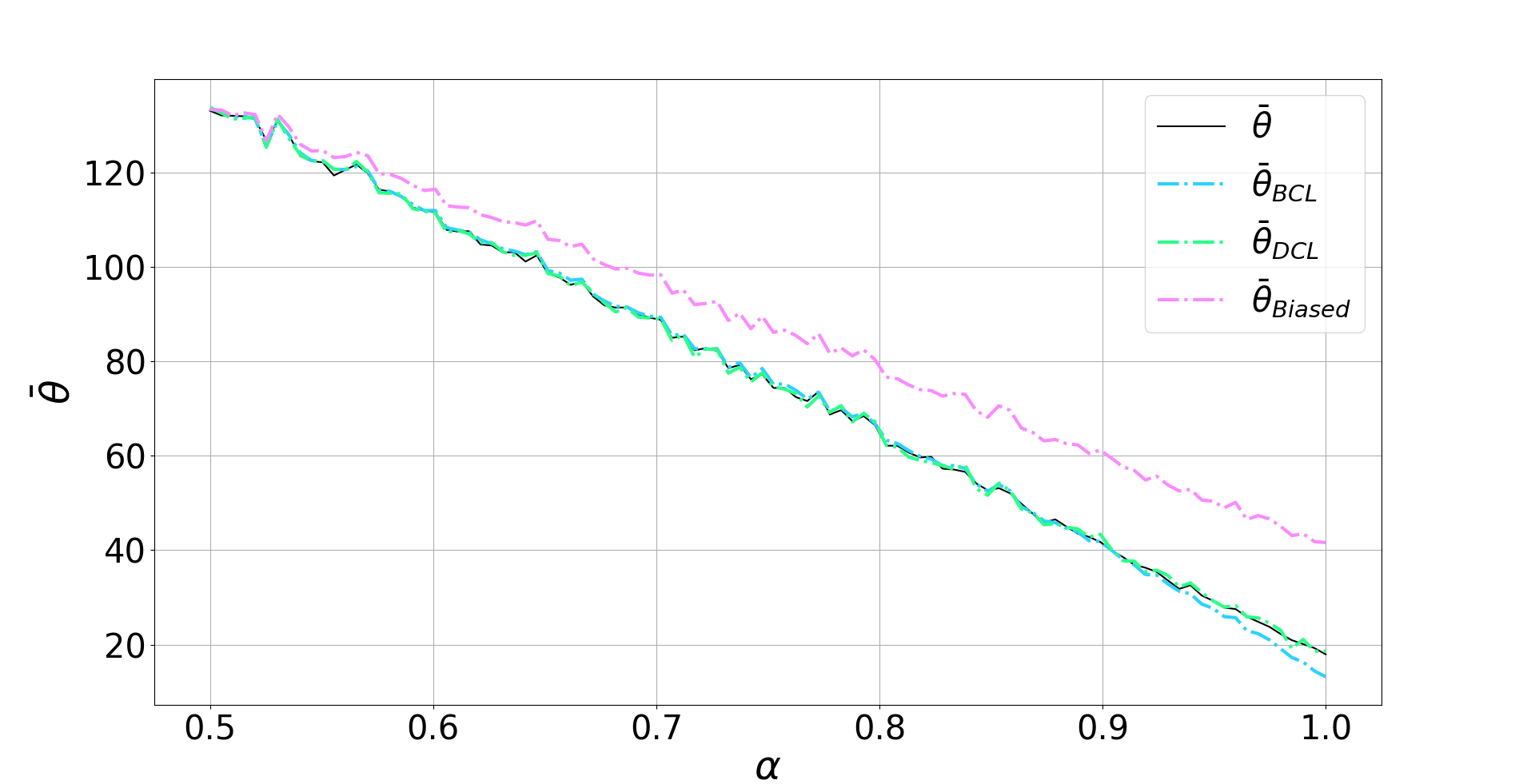}\label{fig:numerbar_alpha}}
	\caption{Mean values of estimators $\hat\theta$ over all anchors.}
	\label{fig:numer_bar}
\end{figure}

\textbf{Mean values}. Aside from $MSE$, we report the mean values of the estimators $\hat\theta $. Our primary concern is whether the chosen of proposal distribution $\phi$ or encoder $f$ affects the consistency of estimators. Fig~\ref{fig: numerbar_gamma} shows the influence of different variation of anchor-specific proposal distribution $\phi$ on mean values of estimators. Fig~\ref{fig:numerbar_alpha} shows the influence of encoder's performance on mean values of estimators. It can be seen that the mean value of $\hat\theta_\textsc{~Bcl}$, $\hat\theta_\textsc{~Dcl}$ and $\theta$ sequences are very close,  which guaranteed by Chebyshev law of large numbers, namely $\lim\limits_{M\rightarrow\infty} \mathbb{P}\{|\frac{1}{M}\sum_{m=1}^{M} \hat{\theta}_m-\frac{1}{M}\sum_{m=1}^{M} \theta_m |< \epsilon \} =1
$ as $\mathbb E ~\hat\theta_m = \theta_m$. This conclusion is un-affected by the chosen of proposal distribution $\phi$ or encoder $f$. Aside from consistency of estimators, Fig~\ref{fig:numerbar_alpha} can be seen as a simulation of training process: a better trained encoder with higher macro-AUC embeds true negative samples dissimilar with anchor, while more similar to anchor for false negative samples. So we observe the decrease of $\bar\theta$, which corresponds to the decrease of loss values in the training process, and increase of bias $|\bar\theta_\textsc{Biased} -\bar\theta|$ since the contained false negative samples in $\bar\theta_\textsc{Biased}$ are scored higher as $\alpha$ increases.

\newpage	
\section{Image Classification}\label{sec:image}
We perform image classification task on 4 public datasets.

\subsection{Data Statistics}\label{appc:data}
\begin{table*}[h!]
	\centering
	\small
	\caption{Data statistics of multi-class image classification}\label{5Table:Dataset}
	\resizebox{0.8\textwidth}{!}{
		\begin{tabular}{lccccc}
			\toprule[1.2pt]
			~           & Number of Classes  & Size  &Training set  &Validation Set&Test Set   \\ \cline{1-6}
			CIFAR10   &   10    &  32x32(RGB)   &50,000&   /	   & 10,000 \\
			STL10    &   10  &  96x96(RGB)  &105,000&  /     & 8,000  \\
			CIFAR100       &   100 &  32x32(RGB)  &50,000 &  /      & 10,000\\
			tinyImageNet       & 200&  64x64(RGB) &  100,000&   10,000&    10,000    \\
			\bottomrule[1.2pt]
		\end{tabular}
	}
\end{table*}

Next, we randomly select two classes from the aforementioned multi-class dataset, resulting in a pseudo-negative class probability of 1/2. Since the STL10 dataset is partially labeled, containing only 5000 labeled samples, we randomly select samples from these labeled samples belonging to the two chosen classes for training. Evaluation is conducted on the test set. Similarly, no labels are utilized during the training process, and labels are solely used for evaluation purposes. The statistical information of the newly constructed dataset is presented in the following table:
\begin{table*}[h!]
	\centering
	\small
	\caption{Data statistics of binary image classification}\label{6Table:Dataset}
	\resizebox{0.8\textwidth}{!}{
		\begin{tabular}{lccccc}
			\toprule[1.2pt]
			~           & Number of Classes  & Size  &Training set  &Validation Set&Test Set   \\ \cline{1-6}
			CIFAR10   &   2    &  32x32(RGB)   &10,000&   /	   & 2,000 \\
			STL10    &   2  &  96x96(RGB)  &1,000&  /     & 1600  \\
			CIFAR100       &   2&  32x32(RGB)  &1,000 &  /      & 200\\
			tinyImageNet       & 2&  64x64(RGB) &  1,000&  /&    100    \\
			\bottomrule[1.2pt]
		\end{tabular}
	}
\end{table*}

\subsection{Implementation Details}\label{appc:deta}
All the experimental settings are identically as DCL~\cite{Chuang:2020:NIPS} and HCL~\cite{Robinson:2021:ICLR}. Specifically, we implement SimCLR~\cite{Chen:2020:ICML} with ResNet-50~\cite{He:2016:CVPR} as the
encoder architecture and use the Adam optimizer~\cite{Adam:2015:ICLR} with learning rate 0.001. The temperature scaling and  the dimension of the latent vector were set as $t$ = 0.5 and $d$=128. All the models are trained for 400 epochs, and evaluated by training a linear classifier after fixing the learned embedding~\cite{Lajanugen:2018:ICLR,Robinson:2021:ICLR}.  Our experimental setup aligns with the existing literature and provides a fair comparison in terms of the considered tasks. A particular note is that no data labels are used during training, and they are only used for evaluation. The source codes are available at \url{https://anonymous.4open.science/r/BCL-2A1B/README.md}

\begin{table*}[!]
	\centering
	\small
	\caption{Impacts of class prior $\tau^+$}\label{6Table:prior}
	\resizebox{0.5\textwidth}{!}{
		\begin{tabular}{lcccc}
			\toprule[1.2pt] 
			Datasets &	Batch Size &$\tau^+$	&	top1&	top5\\ \cline{1-5}
			tinyImageNet&	256	&1e-3&	57.10&	80.54\\
			tinyImageNet&	256	&5e-3&	57.32&	80.87\\
			tinyImageNet&	256 &	1e-2&	56.34&	80.24\\
			\bottomrule[1.2pt]
		\end{tabular}
	}
\end{table*}

\subsection{Impacts of class prior $\tau^+$}\label{appc:para}
The experimental results on the tinyImageNet dataset using different values of $\tau^+$ is presented in Table\ref{6Table:prior}, and the results indicate that the best performance is achieved when $\tau^+$ is set to 1/C, where C is the number of classes. 

To provide some intuition behind this choice, let's consider a specific class, such as "cat", in the dataset. We can view the number of images belonging to the cat class, denoted as X, as the outcome of N independent Bernoulli trials, i.e., X $\sim$ B(N, tau), where N is the number of all samples in the dataset, and tau represents the prior probability of the positive class. In this case, tau can be estimated as $\tau^+= X/N$ , which is an unbiased estimate of the positive class prior. In a balanced dataset, where the total number of samples in the dataset is $N = C * X$, with C being the number of classes, and X being the number of samples in each class, we can calculate that $\tau = 1/C$.

The prior probability of the positive class, $\tau^+$, is determined by the total number of sample classes, C. A higher value of C results in a lower prior probability for the positive class, highlighting the significance of hard negative mining. Conversely, a lower value of C leads to a higher prior probability for the positive class, emphasizing the importance of the task to debias false-negative examples. Therefore, a heuristic approach to setting the parameter beta for hard negative mining is
\[\beta = 1-1/C\]
With a larger total number of classes, indicating a smaller class prior probability, there should be a stronger emphasis on the task of hard negative mining, thereby warranting the selection of a larger $\beta$ value.

\section{Graph Embedding}\label{sec:graph}
We consider the collaborative filtering task. Collaborative filtering tasks require the learning of user and item representations on a user-item bipartite graph, followed by the prediction of personalized rankings based on the similarity between item embeddings and user embeddings. 

\subsection{Experimental Setup}\label{appd:set}
We train the encoders using InfoNCE, DCL, and HCL, respectively. Then, we utilize the learned user-item representations to predict personalized rankings. The performance is evaluated by measuring precision, recall, and NDCG (Normalized Discounted Cumulative Gain) metrics for the top-k recommendation lists. We adopt lightGCN (Light Graph Convolutional Network) and matrix factorization (MF) to encode user and item representations.

\subsection{Data Statistics}\label{appd:data}
The experiments were conducted using 5 public datasets. The collaborative filtering datasets represents a typical positive-unlabeled (PU) dataset, in which only positive feedback, such as ratings, purchases, can be observed. The datasets are divided into training and testing sets with a 4:1 ratio. 
\begin{table*}[h!]
	\centering
	\small
	\caption{Data Statistics}\label{4Table:Dataset}
	\begin{tabular}{lrrrrrr}
		\toprule[1.2pt]
		~           & users  & items  &interactions & training set  &test set&density \\ \cline{1-7}
		MovieLens-100k   &   943    &  1,682   &100,000&    80k	   & 20k &0.06304\\
		MovieLens-1M    &   6,040  &  3,952   &1,000,000&  800k     & 200k&0.04189  \\
		Yahoo!-R3       &   5,400  &  1,000  &182,000 &   146k      & 36k&0.03370\\
		Yelp2018       &   31,668  &  38,048&   1,561,406&   1,249k     & 312k&0.00130  \\
		Gowalla       &   29,858 &  40,981  &1,027,370 &   821k     & 205k&0.00084 \\
		\bottomrule[1.2pt]
	\end{tabular}
\end{table*}
\subsection{Experimental Results}\label{appd:res}
\begin{table*}[h!]
	\centering
	\caption{Top-k ranking performance}\label{5Table:Recommendation}
	\resizebox{1\textwidth}{!}{
		\begin{tabular}{lllccccccccccc}
			\toprule[1.2pt]
			\multirow{2}*{\textbf{Dataset}} & \multirow{2}*{\textbf{Encoder}} & \multirow{2}*{\textbf{Learning Method}} & \multicolumn{3}{c}{Top-5} &~& \multicolumn{3}{c}{Top-10}&~&\multicolumn{3}{c}{Top-20}\\ \cline{4-6} \cline{8-10} \cline{12-14}
			~ & ~ & ~ & Precision& Recall& NDCG& ~ &Precision& Recall& NDCG& ~ &Precision& Recall& NDCG \\ \hline
			
			\multirow{10}*{\textbf{MovieLens-100k}} & \multirow{5}*{\textbf{MF}} &  Info\_NCE &0.4081 & 0.1388 & 0.4324 & ~ & 0.3452 & 0.2266 & 0.4095 & ~ & 0.2793 & 0.3497 & 0.4118 \\
			~ & ~ & DCL &0.4168 & 0.1434 & 0.4458 & ~ & 0.3513 & 0.2291 & 0.4202 & ~ & 0.2835 & 0.3546 & 0.4207 \\ 
			~ & ~ & HCL  & 0.4263 & 0.1463 & 0.4539 & ~ & 0.3565 & 0.2323 & 0.426 & ~ & 0.2849 & 0.3564 & 0.4242 \\	
			~ & ~ &BCL(Proposed)    &0.4374 & 0.1552 & 0.4674 & ~ & 0.3658 & 0.2405 & 0.4380 & ~ & 0.2931 & 0.3588 & 0.4357 \\
			\cline{2-14}
			
			~ & \multirow{5}*{\textbf{LightGCN}}  & Info\_NCE & 0.3924 & 0.1343 & 0.4209 & ~ & 0.3349 & 0.2183 & 0.4006 & ~ & 0.2679 & 0.3289 & 0.3976 \\ 
			~ & ~ & DCL & 0.3962 & 0.1367 & 0.4243 & ~ & 0.3361 & 0.2194 & 0.4022 & ~ & 0.2695 & 0.3329 & 0.4006 \\ 
			~ & ~ & HCL & 0.4197 & 0.1461 & 0.4501 & ~ & 0.3458 & 0.2256 & 0.4188 & ~ & 0.2802 & 0.3446 & 0.4182 \\
			~ & ~ & BCL(proposed) & 0.4366 & 0.1512 & 0.4652 & ~ & 0.3626 & 0.2367 & 0.4351 & ~ & 0.2936 & 0.3599 & 0.4347 \\ \hline\hline

			\multirow{10}*{\textbf{MovieLens-1M}} & \multirow{5}*{\textbf{MF}} & InfoNCE & 0.3820 & 0.0879 & 0.4003 & ~ & 0.3339 & 0.1478 & 0.3728 & ~ & 0.2821 & 0.2358 & 0.3605 \\ 
			~ & ~ & DCL &  0.4009 & 0.0934 & 0.4209 & ~ & 0.3472 & 0.1546 & 0.3894 & ~ & 0.289 & 0.2423 & 0.3731\\ 
			~ & ~ & HCL & 0.4112 & 0.0969 & 0.4317 & ~ & 0.3552 & 0.1585 & 0.3991 & ~ & 0.2959 & 0.2475 & 0.3825 \\ 
			~ & ~ & BCL(proposed) & 0.4249 & 0.1041 & 0.4439 & ~ & 0.3652 & 0.1657 & 0.4101 & ~ & 0.3012 & 0.2537 & 0.3920 \\ 
			\cline{2-14}
			~ & \multirow{5}*{\textbf{LightGCN}} &InfoNCE & 0.4121 & 0.0986 & 0.4386 & ~ & 0.359 & 0.1594 & 0.4041 & ~ & 0.2979 & 0.2482 & 0.3869 \\ 
			~ & ~ & DCL & 0.4104 & 0.0982 & 0.4291 & ~ & 0.3544 & 0.1597 & 0.3977 & ~ & 0.2965 & 0.2511 & 0.3842 \\ 
			~ & ~ & HCL & 0.4107 & 0.0948 & 0.4300 & ~ & 0.3514 & 0.1542 & 0.3950 & ~ & 0.2916 & 0.2413 & 0.3775 \\ 
			~ & ~ & BCL(proposed) & 0.4256 & 0.1047 & 0.4460 & ~ & 0.3651 & 0.1658 & 0.1893 & ~ & 0.2998 & 0.2534& 0.3911 \\\hline \hline
			
			\multirow{10}*{\textbf{Yahoo!-R3}} & \multirow{5}*{\textbf{MF}} & Info\_NCE & 0.1429 & 0.1065 & 0.1615 & ~ & 0.1080 & 0.1601 & 0.1664 & ~ & 0.0786 & 0.2316 & 0.1952 \\ 
			~ & ~ & DCL &0.1454 & 0.1083 & 0.1635 & ~ & 0.1091 & 0.1618 & 0.1692 & ~ & 0.079 & 0.2327 & 0.1974  \\ 
			~ & ~ & HCL & 0.1460 & 0.1097 & 0.1638 & ~ & 0.1096 & 0.1628 & 0.1697 & ~ & 0.0792 & 0.2336 & 0.1976 \\ 
			~ & ~ & BCL(proposed) & 0.1529& 0.1122 & 0.1685 & ~ & 0.1124 & 0.1667 & 0.1736 & ~ & 0.0815 & 0.2367 & 0.2029 \\ 
			\cline{2-14}
			~ & \multirow{5}*{\textbf{LightGCN}} & Info\_NCE & 0.1417 & 0.1074 & 0.1676 & ~ & 0.1099 & 0.1633 & 0.1719 & ~ & 0.0798 & 0.2354 & 0.2007  \\ 
			~ & ~ & DCL & 0.1456 & 0.1092 & 0.1642 & ~ & 0.1089 & 0.1622 & 0.1697 & ~ & 0.079 & 0.2333 & 0.1982\\ 
			~ & ~ & HCL & 0.1412 & 0.1139 & 0.1718 & ~ & 0.113 & 0.1683 & 0.1776 & ~ & 0.0812 & 0.2394 & 0.2059 \\ 
			~ & ~ & BCL(proposed) & 0.1537 & 0.1140 &  0.1725 & ~ & 0.1156 & 0.1696  & 0.1771 & ~ & 0.0842 & 0.2428 & 0.2076\\ \hline\hline

			\multirow{10}*{\textbf{Yelp2018}} & \multirow{5}*{\textbf{MF}} & Info\_NCE  & 0.0429 & 0.0246 & 0.047 & ~ & 0.0365 & 0.0417 & 0.0491 & ~ & 0.0305 & 0.07 & 0.058 \\ 
			~ & ~ & DCL & 0.0486 & 0.0278 & 0.0531 & ~ & 0.0410 & 0.0466 & 0.0552 & ~ & 0.0342 & 0.0777 & 0.0648 \\
			~ & ~ & HCL & 0.0515 & 0.0305 & 0.0566 & ~ & 0.0459 & 0.0541 & 0.0622 & ~ & 0.0383 & 0.0894& 0.0736 \\ 
			~ & ~ & BCL(proposed) & 0.0546 & 0.0342 & 0.0595 & ~ & 0.0475 & 0.0563 & 0.0643 & ~ & 0.0398 & 0.0912 & 0.0752 \\
			\cline{2-14}
			~ & \multirow{5}*{\textbf{LightGCN}} & Info\_NCE & 0.0553 & 0.0329 & 0.0607 & ~ & 0.0473 & 0.0558 & 0.0642 & ~ & 0.0390 & 0.0911 & 0.0754 \\ 
			~ & ~ & DCL & 0.0559 & 0.0331 & 0.0612 & ~ & 0.0472 & 0.0557 & 0.0642 & ~ & 	0.0391 & 0.0914 & 0.0756 \\ 
			~ & ~ & HCL & 0.0563 & 0.0335 & 0.0617 & ~ & 0.0477 & 0.0564 & 0.0648 & ~ & 0.0393 & 0.0920 & 0.0760 \\  
			~ & ~ & BCL(proposed) & 0.0612 & 0.0376 & 0.0670 & ~ & 0.0524 & 0.0629 & 0.0701 & ~ & 0.0414 & 0.1014 & 0.0815 \\\hline\hline

			\multirow{10}*{\textbf{Gowalla}} & \multirow{5}*{\textbf{MF}} & Info\_NCE & 0.0739 &0.0757 & 0.1016 & ~ & 0.0560 & 0.1122 & 0.1076 & ~ & 0.0422 & 0.1650 & 0.1230\\ 
			~ & ~ & DCL & 0.0746 & 0.0769 &0.1023 & ~ & 0.0568 & 0.1147 & 0.1088 & ~ & 0.0426 & 0.1664 & 0.1238 \\ 
			~ & ~ & HCL & 0.0755 & 0.0774 & 0.1035 & ~ & 0.0574 & 0.1151 & 0.1098 & ~ & 0.0432& 0.1693 & 0.1256 \\ 
			~ & ~ & BCL(proposed) & 0.0827 & 0.0837 & 0.1121 & ~ & 0.0639 & 0.1257& 0.1186 & ~ & 0.0485 & 0.1824 & 0.1353 \\
			\cline{2-14}
			~ & \multirow{5}*{\textbf{LightGCN}} & Info\_NCE & 0.0743 & 0.0760 & 0.1022 & ~ & 0.0566 & 0.1132& 0.1084 & ~ & 0.0423 & 0.1649 & 0.1231\\ 
			~ & ~ & DCL & 0.0748 & 0.0763 & 0.1027 & ~ & 0.0569 & 0.1132 & 0.1088 & ~ & 0.0424 & 0.1656 & 0.1236 \\ 
			~ & ~ & HCL & 0.0794 & 0.0804 & 0.1084 & ~ & 0.0608 &0.1199 & 0.1147 & ~ & 0.0453 & 0.1740 & 0.1319 \\ 
			~ & ~ & BCL(proposed) & 0.0863 &  0.0893 & 0.1156 & ~ &  0.0670 &  0.1332 & 0.1248 & ~ & 0.0490 & 0.1937 & 0.1420  \\ \hline
			\bottomrule[1.5pt]
			
		\end{tabular}
	}
\end{table*}

In the context of recommendation scenarios, the prior probability of positive instances can be determined based on the density of the dataset, reflecting the prior probability of users favoring a particular item. Due to the inherent sparsity of recommendation datasets, the prior probability of positive instances tends to be relatively small, thereby highlighting the dominant role of hard negative mining. However, the presence of excessive smoothing in item embeddings often leads to similarity scores that exhibit minimal variations. Consequently, the weight computation in HCL may be compromised. On the other hand, BCL addresses this issue by computing weights using the empirical cumulative distribution function (C.D.F) of unlabeled samples, which remains unaffected by the magnitude of similarity scores and mitigates the impact of excessively smoothed similarity scores.

\end{document}